\documentclass[nohyperref]{article}

\usepackage{microtype}
\usepackage{graphicx}
\usepackage{booktabs} 

\usepackage{hyperref}

\usepackage[accepted]{icml2022}

\usepackage{amsmath}
\usepackage{amssymb}
\usepackage{mathtools}
\usepackage{amsthm}
\usepackage{tcolorbox}
\usepackage{bm}

\usepackage[capitalize,noabbrev]{cleveref}

\usepackage{tikz}
\usetikzlibrary{bayesnet}

\usepackage{dsfont}
\usepackage{graphicx}
\usepackage{booktabs}
\usepackage{caption}
\usepackage{subcaption}
\usepackage{adjustbox}
\usepackage{enumitem}
\usepackage{amsthm}
\usepackage{xcolor}
\usepackage[algo2e,ruled,vlined]{algorithm2e} 

\usepackage{soul}

\theoremstyle{plain}
\newtheorem{theorem}{Theorem}[section]

\theoremstyle{definition}

\theoremstyle{remark}

\newcommand{\ex}{\mathcal{X}}
\newcommand{\exprime}{\mathcal{X}'}
\newcommand{\arr}{R}

\newcommand{\tzero}{\hat{\Theta}^{(0)}}

\newcommand{\sn}{\mathcal{N}(0, \mathrm{I})}

\newcommand{\qlcb}[2]{Q_{\mathrm{LCB}}^{(#1)}(#2)}
\newcommand{\q}[2]{Q^{(#1)}(#2)}
\newcommand{\qlin}[2]{Q_{\mathrm{lin}}^{(#1)}(#2)}
\newcommand{\Why}[1]{\mathcal{Y}^{(#1)}}
\newcommand{\Tzero}[1]{\hat{\Theta}^{(0)}\left(#1\right)}

\definecolor{deepskyblue}{rgb}{0.0, 0.75, 1.0}
\definecolor{dodgerblue}{rgb}{0.12, 0.56, 1.0}
\definecolor{frenchrose}{rgb}{0.96, 0.29, 0.54}
\definecolor{lavender(floral)}{rgb}{0.71, 0.49, 0.86}

\usepackage[textsize=tiny]{todonotes}

\icmltitlerunning{Why So Pessimistic? Offline RL through Ensembles, and Why Their Independence Matters}

\begin{document}

\twocolumn[
\icmltitle{Why So Pessimistic? Estimating Uncertainties for Offline RL through Ensembles, and Why Their Independence Matters}



\icmlsetsymbol{equal}{*}


\begin{center}
   \textbf{
   {\small \color{gray} Seyed} Kamyar {\small \color{gray} Seyed} Ghasemipour
   },
   \textbf{Shixiang Shane Gu},
   \textbf{Ofir Nachum}\\
    Google Research\\
    \texttt{\{kamyar, shanegu, ofirnachum\}@google.com}
\end{center}


\icmlcorrespondingauthor{Firstname1 Lastname1}{first1.last1@xxx.edu}
\icmlcorrespondingauthor{Firstname2 Lastname2}{first2.last2@www.uk}

\icmlkeywords{Machine Learning, ICML}

\vskip 0.3in



\printAffiliationsAndNotice{\icmlEqualContribution} 

\vspace{-2mm}
\begin{abstract}
Motivated by the success of ensembles for uncertainty estimation in supervised learning, we take a renewed look at how ensembles of $Q$-functions can be leveraged as the primary source of pessimism for offline reinforcement learning (RL). We begin by identifying a critical flaw in a popular algorithmic choice used by many ensemble-based RL algorithms, namely the use of shared pessimistic target values when computing each ensemble member's Bellman error. Through theoretical analyses and construction of examples in toy MDPs, we demonstrate that shared pessimistic targets can paradoxically lead to value estimates that are effectively \emph{optimistic}. Given this result, we propose MSG, a practical offline RL algorithm that trains an ensemble of $Q$-functions with independently computed targets based on completely separate networks, and optimizes a policy with respect to the lower confidence bound of predicted action values. Our experiments on the popular D4RL and RL Unplugged offline RL benchmarks demonstrate that on challenging domains such as antmazes, MSG with deep ensembles surpasses highly well-tuned state-of-the-art methods by a wide margin. Additionally, through ablations on benchmarks domains, we verify the critical significance of using independently trained $Q$-functions, and study the role of ensemble size. Finally, as using separate networks per ensemble member can become computationally costly with larger neural network architectures, we investigate whether efficient ensemble approximations developed for supervised learning can be similarly effective, and demonstrate that they do not match the performance and robustness of MSG with separate networks, highlighting the need for new efforts into efficient uncertainty estimation directed at RL.
Our codebase can be found at \url{https://github.com/google-research/google-research/tree/master/jrl}.

\end{abstract}
]




\section{Introduction}

    Offline reinforcement learning (RL), also referred to as batch RL~\citep{lange2012batch}, is a problem setting in which one is provided a dataset of interactions with an environment in the form of a Markov decision process (MDP), and the goal is to learn an effective policy exclusively from this fixed dataset. Offline RL holds the promise of data-efficiency through data reuse, and improved safety due to minimizing the need for policy rollouts. As a result, offline RL has been the subject of significant renewed interest in the machine learning literature \citep{levine2020offline}.
    
    One common approach to offline RL in the model-free setting is to use approximate dynamic programming (ADP) to learn a $Q$-value function via iterative regression to backed-up \emph{target values}. The predominant algorithmic philosophy with most success in ADP-based offline RL is to encourage obtained policies to remain close to the support set of the available offline data. A large variety of methods have been developed for enforcing such constraints, examples of which include regularizing policies with behavior cloning objectives \citep{kumar2019stabilizing, fujimoto2021minimalist}, performing updates only on actions observed inside \citep{peng2019advantage, nair2020awac, wang2020critic, ghasemipour2021emaq} or close to \citep{fujimoto2019off} the offline dataset, and regularizing value functions to underestimate the value of actions not seen in the dataset \citep{wu2019behavior, kumar2020conservative,kostrikov2021offline}.
    
    The need for such regularizers arises from inevitable inaccuracies in value estimation when function approximation, bootstrapping, and off-policy learning -- i.e. The Deadly Triad \citep{van2018deep} -- are involved. In offline RL in particular, such inaccuracies cannot be resolved through additional interactions with the MDP. Thus, remaining close to the offline dataset limits opportunities for catastrophic inaccuracies to arise. However, recent works have argued that the aforementioned constraints can be overly pessimistic, and instead opt for approaches that take into consideration the \emph{uncertainty} about the value function~\citep{buckman2020importance,jin2021pessimism,xie2021bellman}, thus re-focusing the offline RL problem to that of deriving accurate lower confidence bounds (LCB) of $Q$-values.
    
    In the empirical supervised learning literature, \emph{deep ensembles}\footnote{In the deep learning literature, deep ensembles refers to the setting where the same network architectures are trained using the same data and objective functions, with the only difference in ensemble members being the random weight initialization of the networks.} and their more efficient variants have been shown to be the most effective approaches for uncertainty estimation, towards learning calibrated estimates and confidence bounds with modern neural network function approximators~\citep{ovadia2019can}.
    Motivated by this, in our work we take a renewed look into $Q$-ensembles, and study how to leverage them as the primary source of pessimism for offline RL.
    
    In deep RL, a very popular algorithmic choice is to use an ensemble of $Q$-functions to obtain pessimistic value estimates and combat overestimation bias~\citep{fujimoto2018addressing}. Specifically, in the policy evaluation procedure, all $Q$-networks are updated towards a \emph{shared pessimistic temporal difference target}. Similarly in offline RL, in addition to the main offline RL objective that they propose, several existing methods use such $Q$-ensembles~\citep{wu2019behavior,kumar2019stabilizing,agarwal2020optimistic,smit2021pebl,an2021uncertainty,lee2021sunrise,lee2022offline,ghasemipour2021emaq}.
    
    We begin by mathematically characterizing a critical flaw in the aforementioned ensembling procedure.
    Specifically, we demonstrate that using shared pessimistic targets can paradoxically lead to $Q$-estimates which are in fact \emph{optimistic}! We verify our finding by constructing pedagogical toy MDPs. These results demonstrate that the formulation of using shared pessimistic targets is fundamentally ill-formed.
    
    To resolve this problem, we propose \emph{Model Standard-deviation Gradients} (MSG), an ensemble-based offline RL algorithm. In MSG, each $Q$-network is trained \emph{independently}, \emph{without sharing targets}. Crucially, ensembles trained with independent target values will always provide pessimistic value estimates.
    The pessimistic lower-confidence bound (LCB) value estimate -- computed as the mean minus standard deviation of the $Q$-value ensemble -- is then used to update the policy being trained.
    
    Evaluating MSG on the established D4RL \citep{fu2020d4rl} and RL Unplugged \citep{gulcehre2020rl} benchmarks for offline RL, we demonstrate that MSG matches, and in the more challenging domains such as antmazes, significantly exceeds the prior state-of-the-art.
    Additionally, through a series of ablation experiments on benchmark domains, we verify the significance of our theoretical findings, study the role of ensemble size, and highlight the settings in which ensembles provide the most benefit.

    The use of ensembles will inevitably be a computational bottleneck when applying offline RL to domains requiring large neural network models. Hence, as a final analysis, we investigate whether the favorable performance of MSG can be obtained through the use of modern \emph{efficient ensemble} approaches which have been successful in the supervised learning literature~\cite{lee2015m,havasi2020training,wen2020batchensemble,ovadia2019can}.
    We demonstrate that while efficient ensembles are competitive with the state-of-the-art on simpler offline RL benchmark domains, similar to many popular offline RL methods they fail on more challenging tasks, and cannot recover the performance and robustness of MSG using full ensembles with separate neural networks.
    
    Our work highlights some of the unique and often overlooked challenges of ensemble-based uncertainty estimation in offline RL. 
    Given the strong performance of MSG, we hope our work motivates increased focus into efficient and stable ensembling techniques directed at RL, and that it highlights intriguing research questions for the community of neural network uncertainty estimation researchers whom thus far have not employed sequential domains such as offline RL as a testbed for validating modern uncertainty estimation techniques.
\section{Related Work}


Uncertainty estimation is a core component of RL, since an agent only has a limited view into the mechanics of the environment through its available experience data.
Traditionally, uncertainty estimation has been key to developing proper \emph{exploration} strategies such as upper confidence bound (UCB) and Thompson sampling~\citep{lattimore2020bandit}, in which an agent is encouraged to seek out paths where its uncertainty is high.
Offline RL presents an alternative paradigm, where the agent must act conservatively and is thus encouraged to seek out paths where its uncertainty is low~\citep{buckman2020importance}.
In either case, proper and accurate estimation of uncertainties is paramount. To this end, much research has been produced with the aim of devising provably correct uncertainty estimates~\citep{Thomas15HCPE,feng2020accountable,dai2020coindice}, or at least bounds on uncertainty that are good enough for acting exploratorily~\citep{strehl2009reinforcement} or conservatively~\citep{kuzborskij2021confident}. 
However, these approaches require exceedingly simple environment structure, typically either a finite discrete state and action space or linear spaces with linear dynamics and rewards. 

While theoretical guarantees for uncertainty estimation are more limited in practical situations with deep neural network function approximators, a number of works have been able to achieve practical success, for example using deep network analogues for count-based uncertainty~\citep{ostrovski2017count}, Bayesian uncertainty~\citep{ghavamzadeh2016bayesian,yang2020offline}, and bootstrapping~\citep{osband2019deep,kostrikov2020statistical}.
Many of these methods employ ensembles. In fact, in continuous control RL, it is common to use an ensemble of two value functions and use their minimum for computing a target value during Bellman error minimization~\citep{fujimoto2018addressing}. A number of works in offline RL have extended this to propose backing up minimums or lower confidence bound estimates over larger ensembles~\citep{kumar2019stabilizing,wu2019behavior,agarwal2020optimistic,smit2021pebl,lee2021sunrise,lee2022offline,an2021uncertainty}.
In our work, we continue to find that ensembles are extremely useful for acting conservatively, but the manner in which ensembles are used is critical. Specifically our proposed MSG algorithm advocates for using independently learned ensembles, without sharing of target values, and this important design decision is supported by empirical evidence.

The widespread success of ensembles for uncertainty estimation in RL echoes similar findings in supervised deep learning. 
While there exist proposals for more technical approaches to uncertainty estimation~\citep{li2007robust,neal2012bayesian,pawlowski2017implicit}, ensembles have repeatedly been found to perform best empirically~\citep{lee2015m,lakshminarayanan2016simple}.
Much of the active literature on ensembles in supervised learning is concerned with computational efficiency, with various proposals for reducing the compute or memory footprint of training and inference on large ensembles~\citep{wen2020batchensemble,zhu2019binary,havasi2020training}. While these approaches have been able to achieve impressive results in supervised learning, our empirical results suggest that their performance suffers significantly in challenging offline RL settings compared to deep ensembles.

\section{Pessimistic \texorpdfstring{$\bm{Q}$}{Q}-Ensembles: Independent or Shared Targets?}
    \label{sec:independence_matters}
    
    In this section we identify a critical flaw in how ensembles are commonly employed -- in offline as well as online RL -- for obtaining pessimistic value estimates~\citep{wu2019behavior,kumar2019stabilizing,agarwal2020optimistic,smit2021pebl,an2021uncertainty,lee2021sunrise,lee2022offline,ghasemipour2021emaq,an2021uncertainty},
    which can paradoxically lead to an optimism bonus! We begin by mathematically characterizing this problem and presenting a simple fix. Subsequently, we leverage our results to construct pedagogical toy MDPs demonstrating the practical importance of the identified problem and solution.
    
    \subsection{Mathematical Characterization}
        \label{sec:math_char}
        
        We assume access to a dataset $D$ composed of $(s,a,r,s')$ transition tuples from a Markov Decision Process (MDP) determined by a tuple $M = \langle \mathcal{S}, \mathcal{A}, \mathcal{R}, \mathcal{P}, \gamma \rangle$, corresponding to state space, action space, reward function, transitions dynamics, and discount, respectively. As is standard in RL, we do not assume any knowledge of $\mathcal{R},\mathcal{P}$, other than that implicitly provided by the dataset $D$. In this section, for clarity of exposition, we assume that the policies we consider are \emph{deterministic}, and that our MDPs do not have terminal states.
        
        
        We consider $Q$-value ensemble members given by a parameterization $Q_{\theta^i}$, where $i$ indexes into some set $Z$, which is finite in practice but may be infinite or uncountable in theory. We assume $Z$ has an associated probability space
        allowing us to make expectation $\mathds{E}$ or variance $\mathds{V}$ computations over the ensemble members. Given a fixed policy $\pi$, a general dynamic programming based procedure for obtaining pessimistic value estimates is outlined by the iterative regression below:
        \begin{tcolorbox}
        \begin{enumerate}[leftmargin=*]
            \item Initialize $\theta^i$ for all $i\in Z$.
            \item For $t=1,2,\dots$:
            \begin{itemize}
                \item For each $(s,a,r,s')\in D$ and $i\in Z$ compute target values $y^i(r,s',\pi)$.
                \item For each $i\in Z$, update $\theta^i$ to optimize the regression objective
                \begin{equation*}
                    \frac{1}{|D|} \sum_{(s,a,r,s')\in D}(Q_{\theta^i}(s,a) - y^i(r,s',\pi))^2.
                \end{equation*}
            \end{itemize}
            \item Return a pessimistic $Q$-value function $Q_{\mathrm{pessimistic}}$ based on the trained ensemble.
        \end{enumerate}
        \end{tcolorbox}
        A key algorithmic choice in this recipe is where pessimism should be introduced. This can be done by either (a) pessimistically aggregating $Q$-values after training, i.e. inside Step 3, or (b) \emph{also incorporating pessimism during Step 2,} by using a shared pessimistic target value $y$.
        Through our review of the offline RL (as well as online RL) literature, we have observed that the most common approach is the latter, where the targets are \emph{pessimistic, shared, and identical} across ensemble members~\citep{wu2019behavior,kumar2019stabilizing,agarwal2020optimistic,smit2021pebl,an2021uncertainty,lee2021sunrise,lee2022offline,ghasemipour2021emaq}. Specifically, they are computed as,
        \begin{align*}
            y^i(r, s', \pi) = \mathrm{PO}(\{r + \gamma Q_{\theta^i}(s', \pi(s')), \forall i \in Z\})
        \end{align*}
        with $\mathrm{PO}$ being a desired pessimism operator aggregating the TD target values of the ensemble members. Example pessimism operators include ``mean minus standard deviation", or ``minimum".
        
        In this section, our goal is to compare these two alternative approaches. For our analysis, we will use ``mean minus standard deviation" (a lower confidence bound (LCB)) as our pessimism operator, and use the notation $Q_{\mathrm{LCB}}$ in place of $Q_{\mathrm{pessimistic}}$ (defined in the algorithm box above).
        Under the LCB pessimism operator we will have:
        \begin{itemize}[leftmargin=2mm,label={},nosep]
             \item \textbf{Independent Targets (Method 1):}
             \begin{align*}
             y^i(r, s', \pi) &= r + \gamma \cdot Q_{\theta^i}(s', \pi(s')) \\
             Q_{\mathrm{LCB}}(s,a) &= \mathds{E}\left[Q_{\theta^i}(s, a)\right] - \sqrt{\mathds{V}\left[Q_{\theta^i}(s, a)\right]}
             \end{align*}
             \item \textbf{Shared Targets (Method 2):}
             \begin{align*}
                 y^i(r, s', \pi) &= r + \gamma \cdot \Big(\mathds{E}\left[Q_{\theta^i}(s', \pi(s'))\right] \\& \hspace{16mm} - \sqrt{\mathds{V}\left[Q_{\theta^i}(s', \pi(s'))\right]}\Big) \\
                Q_{\mathrm{LCB}}(s,a) &= \mathds{E}\left[Q_{\theta^i}(s, a)\right] - \sqrt{\mathds{V}\left[Q_{\theta^i}(s, a)\right]}
             \end{align*}
        \end{itemize}

        To characterize the form of $Q_{\mathrm{LCB}}$ 
        when using complex neural networks, we refer to the work on infinite-width neural networks, namely the Neural Tangent Kernel (NTK) ~\citep{jacot2018neural}.
        We consider $Q$-value ensemble members, $Q_{\theta^i}$, which all share the same infinite-width neural network architecture (and thus the same NTK parameterization).
        As noted in the algorithm box above, and as is the case in deep ensembles~\citep{lakshminarayanan2016simple}, the only difference amongst ensemble members $Q_{\theta^i}$ is in their initial weights $\theta^i$ sampled from the neural network's initial weight distribution.
        
        Before presenting our results, we establish some notation relevant to the infinite-width and NTK regime. Let $\mathcal{X}, \arr, \ex'$ denote data matrices containing $(s,a)$, $r$, and $(s',\pi(s'))$ appearing in the offline dataset $D$; i.e., the $k$-th transition $(s,a,r,s')$ in $D$ is represented by the $k$-th rows in $\mathcal{X},\arr,\ex'$.
        Let $A, B$ denote two data matrices, where similar to $\ex, \exprime$, each row contains a state-action tuple $(s, a) \in \mathcal{S} \times \mathcal{A}$. The NTK, which governs the training dynamics of the infinitely-wide neural network, is then given by the outer product of gradients of the neural network \emph{at initialization}:
            \begin{equation}
                \tzero_i(A, B) := \nabla_\theta Q_{\theta^i}(A)  \cdot \nabla_\theta Q_{\theta^i}(B)^T \vert_{t=0}
            \end{equation}
        where we overload notation $Q_{\theta^i}(A)$ to represent the column vector containing $Q$-values.
        At infinite-width in the NTK regime, $\tzero_i(A, B)$ converges to a deterministic kernel (i.e. does not depend on the random weight sample $\theta^i$), and hence is the same for all ensemble members. Thus, hereafter we will remove the index $i$ from the notation of the NTK kernel and simply write, $\tzero(A, B)$.
        With our notation in place, we define,
        \begin{align*}
                C := \tzero(\exprime, \ex) \cdot \tzero(\mathcal{X}, \mathcal{X})^{-1}
                \label{eq:kernel}
        \end{align*}
        Intuitively, $C$ is a $\vert D \vert \times \vert D \vert$ matrix where
        the element at column $q$, row $p$, captures a notion of similarity between $(s, a)$ in the $q^{th}$ row of $\ex$, and $(s', \pi(s'))$ in the $p^{th}$ row of $\exprime$. 
        
        We now have all the necessary machinery to characterize the form of $Q_{\mathrm{LCB}}$ when using either independent or shared targets:

        \begin{theorem}
            \label{theorem:modulation}
            
            For a given $(s,a) \in \mathcal{S} \times \mathcal{A}$, let $Q^{(0)}_{\theta^i}(s,a)$ denote $Q_{\theta^i}(s,a)\vert_{t=0}$ (value at initialization), with $\theta$ sampled from the initial weight distribution.
            After $t+1$ iterations of pessimistic policy evaluation, the LCB value estimate for $(s',\pi(s')) \in \exprime$ is given by,
                \begin{align}
                    &\textbf{Independent Targets (Method 1):} \nonumber\\ &Q_{\mathrm{LCB}}^{(t+1)}(\exprime) = \mathcal{O}(\gamma^t \|C\|^t) + \underbrace{(1 + \ldots + \gamma^{t} C^{t})}_{\text{backup term}}CR \\&-\sqrt{\mathds{E} \Big[\Big( \underbrace{(1 + \ldots + \gamma^{t} C^{t})}_{\text{backup term}} (Q^{(0)}_{\theta^i}(\exprime) - CQ^{(0)}_{\theta^i}(\ex))\Big)^2\Big]} \nonumber\\
                    &\textbf{Shared Targets (Method 2):} \nonumber\\ &Q_{\mathrm{LCB}}^{(t+1)}(\exprime) = \mathcal{O}(\gamma^t \|C\|^t) + \underbrace{(1 + \ldots + \gamma^{t} C^{t})}_{\text{backup term}}C \arr \label{eq:method_2}\\&- \underbrace{(1 + \ldots + \gamma^{t} C^{t})}_{\text{backup term}} \sqrt{\mathds{E}\Big[\Big(Q^{(0)}_{\theta^i}(\exprime) - C Q^{(0)}_{\theta^i}(\ex)\Big)^2\Big]} \nonumber,
                \end{align}
            where the square and square-root operations are applied element-wise.\footnote{Note that if $\gamma\|C\| \ge 1$, dynamic programming is liable to diverge in either setting. In our discussions, we avoid this degenerate case and assume $\gamma\|C\| < 1$.}
            \begin{proof}
                Please refer to Appendix \ref{app:proof}.
            \end{proof}
            
        \end{theorem}

        As can be seen, the equations for the pessimistic LCB value estimates in both settings are similar,
        only differing in the third term. The first term is negligible and tends towards zero as the number of iterations of policy evaluation increases.
        The second term shared by both variants corresponds to the expected result of the
        policy evaluation
        procedure \emph{without any pessimism} (as before, we mean expectation under $\theta$ sampled from the initial weight distribution). Accordingly, the differing third term in each variant exactly corresponds to the ``pessimism'' or ``penalty'' induced by that variant.
        
        Considering the available offline RL dataset $D$ as a restricted MDP in itself, we see that the use of Independent Targets (Method 1) leads to a pessimism term
        that performs ``backups"
        along
        the trajectories that the policy would experience in this restricted MDP
        (using the geometric term $1+\dots+\gamma^{t} C^{t}$) before computing a variance estimate. Meanwhile the use of Shared Targets (Method 2) does the reverse -- it first computes a variance term and then performs the ``backups".
        
        
        While this difference may seem inconsequential, it becomes critical when one realizes that in Equation \ref{eq:method_2} for Shared Targets (Method 2), the pessimism term (third term) may become positive, i.e. a \emph{negative penalty}, yielding an effectively \emph{optimistic} LCB estimate.  Critically, with Independent Targets (Method 1), this problem \emph{cannot occur}.

    \subsection{Validating Theoretical Predictions}
    In this section we demonstrate that our analysis is not solely a theoretical result concerning the idiosyncracies of infinite-width neural networks, but that it is rather straightforward to construct combinations of an MDP, offline data, and a policy, that lead to the critical flaw of an optimistic LCB estimate.
    
    Let $d_s, d_a$ denote the dimensionality of state and action vectors respectively. We consider an MDP whose initial state distribution is a spherical multivariate normal distribution $\sn$, and whose transition function is given by $\mathcal{P}(s'\vert s,a) = \sn$.
    Consider the procedure for generating our offline data matrices, described in the box below. This procedure returns data matrices $\ex, \arr, \exprime$ by generating $N$ episodes of length $T$, using a behavior policy $a \sim \sn$. In this generation process, we set the policy we seek to pessimistically evaluate, $\pi$, to always apply the behavior policy's action in state $s$ to the next state $s'$.
    
    To construct our examples, we consider the setting where we use linear models to represent $Q_{\theta^i}$, with the initial weight distribution being a spherical multivariate normal distribution, $\sn$. With linear models,
    the equations for $Q_\mathrm{LCB}$ takes an identical form to those in Theorem \ref{theorem:modulation}.
    
    \begin{figure}[h]
        \begin{tcolorbox}
            \begin{enumerate}[leftmargin=*]
                \item Initialize empty $\ex, \arr, \exprime$
                \item For $N$ episodes:
                \begin{itemize}
                    \item sample $s \sim \sn$
                    \item For $T$ steps:
                    \begin{itemize}
                        \item sample $a \sim \sn$
                        \item sample $s' \sim \sn$
                        \item set $\pi(s') \leftarrow a$
                        \item Add $(s, a)$ to $\ex$
                        \item Add $r \sim \sn$ to $\arr$
                        \item Add $(s', \pi(s'))$ to $\exprime$
                        \item Set $s \leftarrow s'$
                    \end{itemize}
                \end{itemize}
                \item Return the offline dataset $\ex, \arr, \exprime$
            \end{enumerate}
        \end{tcolorbox}
    \end{figure}
    
    Given the described data generating process and our choice of linear function approximation, we can compute the pessimism term for the Shared Targets (Method 2) in Theorem \ref{theorem:modulation}, Equation \ref{eq:method_2}, namely the term,
    \begin{align}
        - (1 + \ldots + \gamma^{t}C^{t}) \sqrt{\mathds{E}\Big[\Big(Q^{(0)}_{\theta^i}(\exprime) - C Q^{(0)}_{\theta^i}(\ex)\Big)^2\Big]}
        \label{eq:pessimism_term}
    \end{align}
    We implement this computation in a simple Python script, which we include in the supplementary material. We choose, $d_s = 30$, $d_a = 30$, $\gamma = 0.5$, $N = 5$, $T = 5$, and $t = 1000$ ($t$ is the exponent in the geometric term above). We run this simulation $1000$ times, each with a different random seed. After filtering simulation runs to ensure $\gamma\|C\| < 1$ (as discussed in an earlier footnote), we observe that $221$ of the simulation runs result in an optimistic LCB bonus, meaning that in those experiments, the pessimism term in Equation \ref{eq:pessimism_term} above was in fact positive for some $(s', \pi(s')) \in \exprime$. We have made the python notebook implementing this experiment available in our opensourced codebase.
    
    For intriguing investigations in pedagogical toy MDPs regarding the structure of uncertainties, we strongly encourage the interested reader to refer to Appendix \ref{app:old_toy_mdps}.

\section{Model Standard-deviation Gradients (MSG)}
    It is important to note that even if the pessimism term does not become positive for a particular combination of MDPs, offline datasets, and policies, the fact that it can occur highlights that the formulation of Shared Targets is fundamentally ill-formed.
    To resolve this problem we propose Model Standard-deviation Gradients (MSG), an offline RL algorithm which leverages ensembles to approximate the LCB using the approach of Independent Targets.
    
    \subsection{Policy Evaluation and Optimization in MSG}
    
    MSG follows an actor-critic setup. At the beginning of training, we create an ensemble of $N$ Q-functions by taking $N$ samples from the initial weight distribution. During training, in each iteration, we first perform policy evaluation by estimating the $Q_{\mathrm{LCB}}$ for the current policy, and subsequently optimize the policy through gradient ascent on $Q_{\mathrm{LCB}}$.
        
        \vspace{-8pt}
        \paragraph{Policy Evaluation}
        As motivated by our analysis in Section \ref{sec:independence_matters}, we train the ensemble $Q$-functions independently using the standard least-squares Q-evaluation loss,
            \begin{align}
                &\mathcal{L}(\theta^i) = \mathds{E}_{(s,a,r,s') \sim D}\Big[\left(Q_{\theta^i}(s,a) - y^i(r, s',\pi)\right)^2\Big]
                \label{eq:td}\\
                &y^i = r + \gamma \cdot \mathds{E}_{a' \sim \pi(s')}\Big[Q_{\Bar{\theta}^i}(s', a')\Big], \label{eq:target}
            \end{align}
        where $\theta^i, \Bar{\theta}^i$ denote the parameters and target network parameters for the $i^{\text{th}}$ Q-function.
        
        In each iteration, as is common practice, we do not update the $Q$-functions until convergence, and instead update the networks using a single gradient step.
        In practice, the expectation in equation \ref{eq:td} is estimated by a minibatch, and the expectation in equation \ref{eq:target} is estimated with a single action sample from the policy. After every update to the Q-function parameters, their corresponding target parameters are updated to be an exponential moving average of the parameters in the standard fashion.
        
        \vspace{-8pt}
        \paragraph{Policy Optimization}
        As in standard deep actor-critic algorithms, policy evaluation steps (learning $Q$) are interleaved with policy optimization steps (learning $\pi$).
        In MSG, we optimize the policy through gradient ascent on $Q_{\mathrm{LCB}}$. Specifically, our proposed policy optimization objective in MSG is,
            \begin{align}
                \mathcal{L}(\pi) &= \mathds{E}_{s\sim D, a\sim\pi(s)}\left[Q_{\mathrm{LCB}}(s,a)\right] \label{eq:msg_policy_objective}\\
                &=\mathds{E}_{s\sim D, a\sim\pi(s)}\left[\mathds{E}[Q_{\theta^i}(s,a)] + \beta \sqrt{\mathds{V}[Q_{\theta^i}(s,a)]}\right]\nonumber
            \end{align}
        where $\beta \le 0$ is a hyperparameter that determines the amount of pessimism.
    \subsection{The Trade-Off Between Trust and Pessimism}
        \label{sec:tradeoff}
        While our hope is to leverage the implicit generalization capabilities of neural networks to estimate proper LCBs beyond states and actions in the finite dataset $D$, neural network architectures can be fundamentally biased, or we can simply be in a setting with insufficient data coverage, such that the generalization capability of those networks is limited.
        To this end, we augment the policy evaluation objective of MSG (equation~\ref{eq:td}) with a support constraint regularizer inspired by CQL \citep{kumar2020conservative}\footnote{Instead of a CQL-style value regularizer, other forms of support constraints such as a behavioral cloning regularizer on the policy could potentially be used.}:
        \begin{equation*}
            \mathcal{H}(\theta^i) = \mathds{E}_{s \sim D, a\sim\pi(s)}\left[ Q_{\theta^i}(s,a)\right] - \mathds{E}_{(s,a) \sim D}\left[ Q_{\theta^i}(s,a)\right].
        \end{equation*}
        This regularizer encourages the $Q$-functions to increase the values for actions seen in the dataset $D$, while decreasing the values of the actions of the current policy.
        Practically, we estimate the latter expectation of $\mathcal{H}$ using the states in the mini-batch, and we approximate the former expectation using a single sample from the policy.
        We control the contribution of $\mathcal{H}(\theta^i)$ by weighting this term with weight parameter $\alpha$. The full critic loss is thus given by,
        \begin{equation}
            \mathcal{L}(\theta^1,\dots,\theta^N) = \sum_{i=1}^N \left(\mathcal{L}(\theta^i) + \alpha \mathcal{H}(\theta^i)\right) \label{eq:full_msg_q_objective}
        \end{equation}
        Empirically, as evidenced by our results in Appendix \ref{app:gym_baselines}, we have observed that such a regularizer can be necessary in two situations: 1) The first scenario is where the offline dataset only contains a narrow data distribution (e.g., imitation learning datasets only containing expert data). We believe this is because the power of ensembles comes from predicting a value distribution for unseen $(s,a)$ based on the available training data. Thus, if no data for sub-optimal actions is present, ensembles cannot make accurate predictions and increased pessimism
        via $\mathcal{H}$ becomes necessary. 2) The second scenario is where environment dynamics can be chaotic (e.g. Gym~\citep{brockman2016openai} \texttt{hopper} and \texttt{walker2d}). In such domains it would be beneficial to remain close to the observed data in the offline dataset.
        
        Pseudo-code for our proposed MSG algorithm can be viewed in Algorithm Box \ref{alg:full_alg}.


\section{Experiments}
\begin{figure}[t]
    \centering
    \vspace{3mm}
    \includegraphics[width=0.3\textwidth]{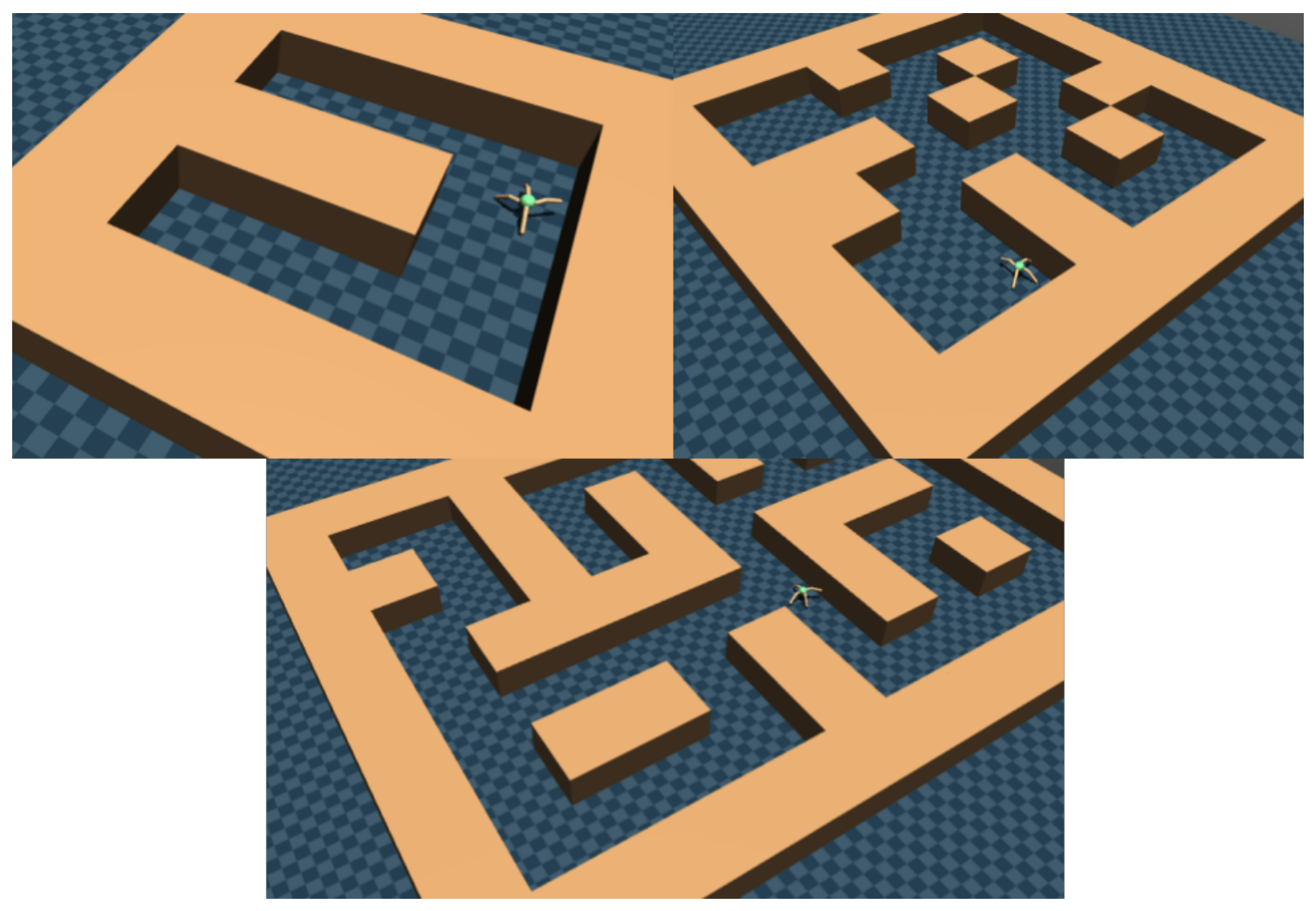}
    \caption{\small{
        D4RL antmaze tasks. Figure taken from \citet{fu2020d4rl}.
    }}
    \vspace{-3mm}
    \label{fig:my_label}
\end{figure}


\vspace{-6pt}
\begin{figure*}[t]
    \centering
    \begin{adjustbox}{center}
    \includegraphics[width=\textwidth]{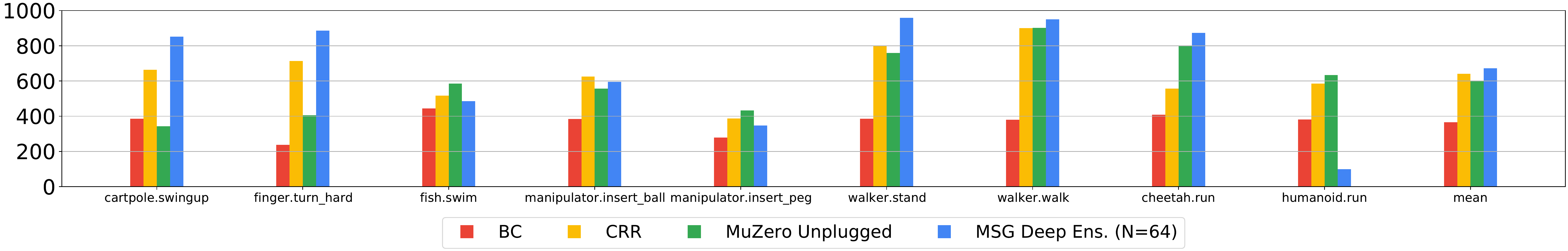}
    \end{adjustbox}
    \caption{
        \small{
            Results for DM Control Suite subset of the RL Unplugged benchmark \citep{gulcehre2020rl}.
            We note that: 1) the architecture used for MSG is smaller by a factor of approximately 60x which contributes to poor performance on \texttt{humanoid.run}, 2) CRR results are reported by their best checkpoint throughout training which differs from MSG, BC, and MuZero Unplugged which report performance at the end of training. Baseline results taken from \citet{schrittwieser2021online}.
            Despite MSG's disadvantage on the \texttt{humanoid.run} task, it still edges out the baseline methods in mean performance.
        }
    }
    \label{fig:rlu}
\end{figure*}

\begin{table*}[ht]
    \centering
    \small{
        \begin{tabular}{|l|c|c|ccc|ccc|}
        \toprule
        {Domain} &  CQL (Reported)  &   IQL (Reported)      &       MSG ($N=64$)    &   $\beta$    &   $\alpha$  &   MSG ($N=4$)    &   $\beta$    &   $\alpha$\\
        \midrule
        maze2d-umaze-v1 &   5.7   &  -- &  $\bm{101.1 \pm 26.3}$   &   $-8$    &  $0$   & $68.8 \pm 20.2$ & $-4$ & $0.1$  \\
        maze2d-medium-v1 &  5.0 & -- &   $\bm{57.0 \pm 17.2}$     &     $-4$  &    $0.1$     & $\bm{53.2 \pm 26.8}$ & $-4$ & $0.1$ \\
        maze2d-large-v1 &  12.5 & --   &   $\bm{159.3 \pm  49.4}$    &  $-4$  &  $0.1$   & $59.2 \pm 59.1$ & $-8$ & $0.5$ \\
        \midrule
        antmaze-umaze-v0 &  74.0 & 87.5   & $\bm{97.8 \pm 1.2}$ & $-4$ & $0.5$ & $\bm{98.6 \pm 1.4}$ & $-4$ & $1.0$ \\
        antmaze-umaze-diverse-v0 & $\bm{84.0} $ & 62.2 & $\bm{81.8 \pm 3.0}$ & $-4$ & $1.0$ &   $76.6 \pm 7.6$ &       $-4$        &        $0.5$        \\
        antmaze-medium-play-v0 &  61.2 &  71.2  & $\bm{89.6 \pm 2.2}$ & $-4$ & $0.5$ &   $\bm{83.0 \pm 7.1}$          & $-4$ & $0.1$ \\
        antmaze-medium-diverse-v0 & 53.7 & 70.0 & $\bm{88.6 \pm 2.6}$ & $-4$ & $0.5$ &  $\bm{83.0 \pm 6.2}$ & $-4$ & $0.5$ \\
        antmaze-large-play-v0 &15.8 & 39.6 & $\bm{72.6 \pm 7.0}$ & $-8$ & $0$ &   $46.8 \pm 14.7$ & $-4$ & $0.5$ \\
        antmaze-large-diverse-v0 &   14.9 & 47.5 & $\bm{71.4 \pm 12.2}$ & $-8$ & $0.1$ &      $58.2 \pm 9.6$           & $-8$ & $0.1$\\
        \bottomrule
        \end{tabular}
    }
    \caption{
        \small{
            Results on D4RL maze2d and antmaze domains. In MSG, $\beta$ is the hyperparameter controlling the amount of pessimism in $Q_{LCB}$ (Equation \ref{eq:msg_policy_objective}), and $\alpha$ is the hyperparameter controlling the contribution of the CQL-style regularizer (Equation \ref{eq:full_msg_q_objective}). As we were unable to reproduce CQL antmaze results despite extensive hyperparameter tuning (see also~\citet{kostrikov2021offline}), we present the numbers reported by the original paper which uses the same network architectures as MSG.
        }
    }
    \label{tab:hard_tasks}
\end{table*}


    
    In this section we seek to empirically answer the following questions: 1) How well does MSG perform compared to current state-of-the-art in offline RL? 2) Are the theoretical differences in ensembling approaches (Section \ref{sec:independence_matters}) practically relevant? 3) When and how does ensemble size affect perfomance? 4) Can we match the performance of MSG through efficient ensemble approximations developed in the supervised learning literature?
    
    
    \subsection{Offline RL Benchmarks}
        
        
        \paragraph{D4RL Gym Domains}
        \label{sec:gym_d4rl}
        We begin by evaluating MSG on the Gym domains (\texttt{halfcheetah}, \texttt{hopper}, \texttt{walker2d}) of the D4RL offline RL benchmark \citep{fu2020d4rl}, using the \texttt{medium}, \texttt{medium-replay}, \texttt{medium-expert}, and \texttt{expert} data settings. Our results presented in Appendix \ref{app:gym_baselines} (summarized in Figure \ref{fig:summarrized_bar_plots}) demonstrates that MSG is competitive with well-tuned state-of-the-art methods CQL~\citep{kumar2020conservative} and F-BRC~\citep{kostrikov2021offline}.
        
        \paragraph{D4RL Antmaze Domains}
        \label{sec:d4rl_antmaze}
        Due to the narrow range of behaviors in Gym environments, offline datasets for these domains tend to be very similar to imitation learning datasets. As a result, many prior offline RL approaches that perform well on D4RL Gym fail on harder tasks that require stitching trajectories through dynamic programming (c.f. \citet{kostrikov2021offlineIQL}).
        An example of such tasks are the D4RL antmaze settings, in particular those in the \texttt{antmaze-medium} and \texttt{antmaze-large} environments. The data for antmaze tasks consists of many episodes of an Ant agent \citep{brockman2016openai} running along arbitrary paths in a maze. The agent is tasked with using this data to learn a point-to-point navigation policy from one corner of the maze to the opposite corner, where rewards are given by a sparse signal that is $1$ when near the desired end location in the maze -- at which point the episode is terminated -- and $0$ otherwise.
        The undirected, extremely sparse reward nature of antmaze tasks make them very challenging, especially for the large maze sizes.
        Table \ref{tab:hard_tasks} and Appendix \ref{app:antmaze_baselines} present our results.
        To the best of our knowledge, the antmaze domains are considered unsolved, with few prior works reporting non-zero results on the large mazes~\citep{kumar2020conservative,kostrikov2021offlineIQL}.
        As can be seen, MSG obtains results that far exceed the prior state-of-the-art results reported by \citet{kostrikov2021offlineIQL}.
        While some works that use specialized hierarchical approaches have reported strong results as well~\citep{ajay2020opal},
        it is notable that MSG is able to solve these challenging tasks with standard architectures and training procedures, and this shows the power that ensembling can provide -- as long as the ensembling is performed properly!
        
        
        

    \paragraph{RL Unplugged}
        
        In addition to the D4RL benchmark,
        we evaluate MSG on the RL Unplugged benchmark~\citep{gulcehre2020rl}. 
        Our results are presented in Figure \ref{fig:rlu}. 
        We compare to results for Behavioral Cloning (BC) and two state-of-the-art methods in these domains, Critic-Regularized Regression (CRR) \citep{wang2020critic} and MuZero Unplugged \citep{schrittwieser2021online}.
        Due to computational constraints when using deep ensembles, we use the same network architectures as we used for D4RL experiments. The networks we use are approximately $\frac{1}{60}$-th the size of those used by the BC, CRR, and MuZero Unplugged baselines in terms of number of parameters.
        We observe that MSG is on par with or exceeds the current state-of-the-art on all tasks with the exception of \texttt{humanoid.run},
        which appears to require the larger architectures used by the baseline methods.
        Additional experimental details can be found in Appendix \ref{app:rlu}.
        
    \paragraph{Benchmark Conclusion} Prior work has demonstrated that many offline RL approaches that perform well on Gym domains, fail to succeed on much more challenging domains \citep{kostrikov2021offlineIQL}. Our results demonstrate that through uncertainty estimation with deep ensembles, MSG is able to very significantly outperform prior work on very challenging benchmark domains such as the D4RL antmazes.
        

\subsection{Ensemble Ablations}
    \label{sec:ablations}
    \paragraph{Independence in Ensembles}
    In Section \ref{sec:independence_matters}, through theoretical arguments and toy experiments we demonstrated the importance of training using ``Independent" ensembles. Here, we seek to validate the significance of our theoretical findings using offline RL benchmarks, by comparing \texttt{Independent} targets (as in MSG), to \texttt{Shared-LCB} and \texttt{Shared-Min} targets. Our results are presented in Appendices \ref{app:gym_ablations} and \ref{app:antmaze_ablations}, with a summary in Figure \ref{fig:summarrized_bar_plots}.
    
    In the Gym domains (Appendix \ref{app:gym_ablations}), with ensemble size $N=4$, \texttt{Shared-LCB} significantly underperforms MSG. In fact, not using ensembles at all ($N=1$) outperforms \texttt{Shared-LCB}. With ensemble size $N=4$, \texttt{Shared-Min} is on par with MSG. When the ensemble size is increased to $N=64$ (Figure \ref{fig:gym_msg_vs_shared_min}), we observe the performance of \texttt{Shared-Min} drops significantly on $7 / 12$ D4RL Gym settings. In constrast, the performance of MSG is stable and does not change.
    
    In the challenging antmaze domains (Appendix \ref{app:antmaze_ablations}), for both ensemble sizes $N=4$ and $N=64$, \texttt{Shared-LCB} and \texttt{Shared-Min} targets completely fail to solve the tasks, while for both ensemble sizes MSG exceeds the prior state-of-the-art (Table \ref{tab:hard_tasks}), IQL \citep{kostrikov2021offlineIQL}.
    
    \paragraph{Independence in Ensembles Conclusion} Our experiments corroborate the theoretical results in Section \ref{sec:independence_matters}, demonstrating that \texttt{Independent} targets are critical to the success of MSG. These results are particularly striking when one considers that the implementations for MSG, \texttt{Shared-LCB}, and \texttt{Shared-Min} differ by only 2 lines of code.
    
    \paragraph{Ensemble Size}
    \begin{figure}[t]
        \centering
        \includegraphics[width=0.5\textwidth]{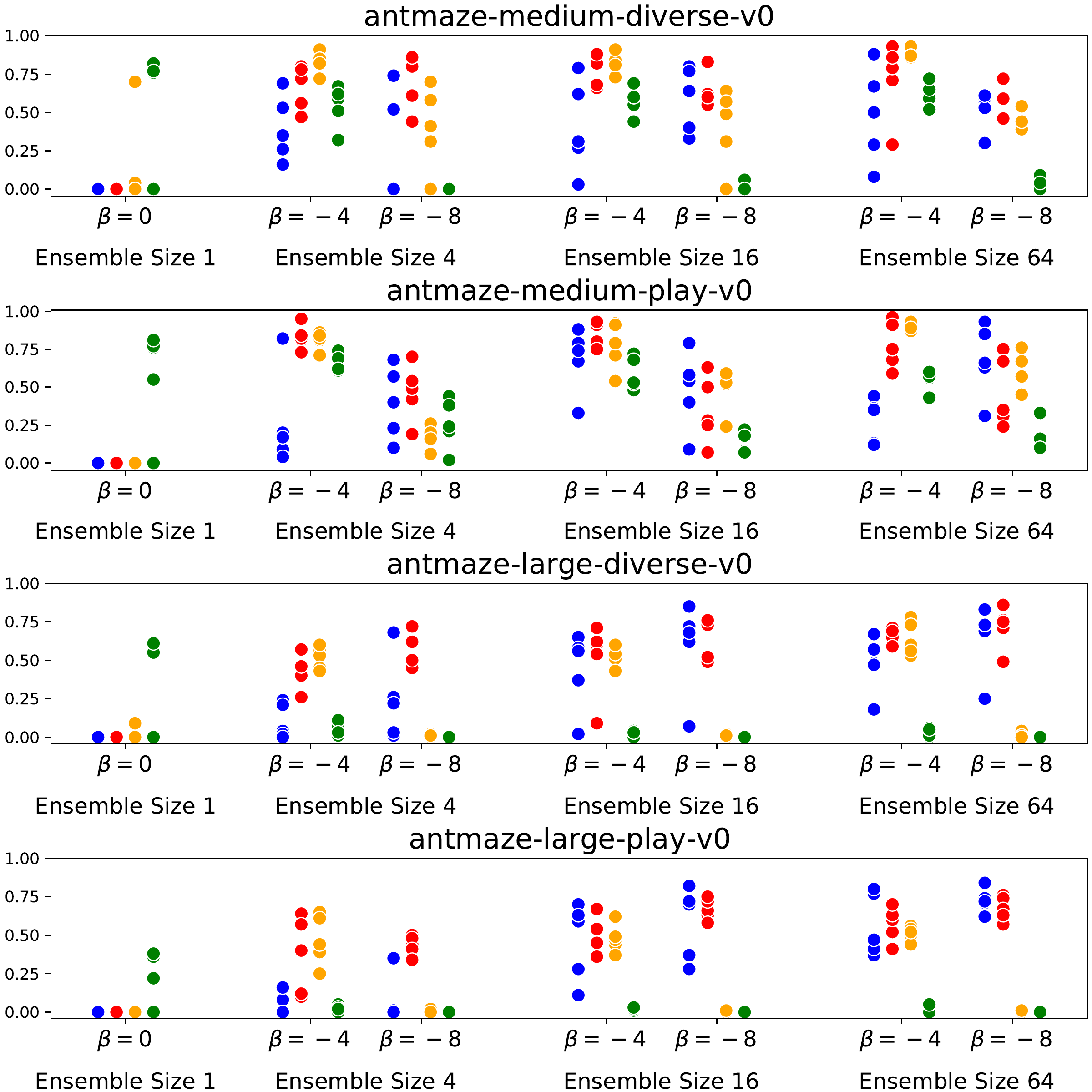}
        \caption{
            \small{
                Ensemble size ablation on the antmaze medium and large domains.
                We compare ensemble size $\{1, 4, 16, 64\}$. For each task, for each ensemble size, we ran 5 experiments for each combination of $\beta \in \{-4, -8\}$ and $\alpha \in \{0, 0.1, 0.5, 1\}$ (colored blue, red, yellow, and green respectively). For ensemble size 1 we use $\beta = 0$ as the mean and LCB are equivalent. We observe the general trend that bigger ensembles lead to better performance, despite variances across hyperparameters.
            }
        }
        \label{fig:ens_size_no_sass}
    \end{figure}
    Another important ablation is to understand the role of ensemble size in MSG.
    
    In the Gym domains, Figure \ref{fig:gym_baselines} demonstrates that increasing the number of ensembles from $4$ to $64$ does not result in a noticeable change in performance.
    
    In the antmaze domains, we evaluate MSG under ensemble sizes $\{1, 4, 16, 64\}$. Figure \ref{fig:ens_size_no_sass} presents our results. Our key takeaways are as follows:
    \begin{itemize}
        \item For the harder \texttt{antmaze-large} tasks, there is a clear upward trend as ensemble size increases.
        \item Using a small ensemble size (e.g. $N=4$) is already quite good, but more sensitive to hyperparameter choice especially on the harder tasks.
        \item Very small ensemble sizes benefit more from using $\alpha > 0$ \footnote{As a reminder, $\alpha$ is the weight of the CQL-style regularizer loss discussed in Section \ref{sec:tradeoff}.}. However, across the board, using $\alpha=0$ is preferable to using too large of a value for $\alpha$ -- with the exception of $N=1$ which cannot take advantange of the benefits of ensembling.
        \item When using lower values of $\beta$, lower values of $\alpha$ should be used.
    \end{itemize}
    
    \paragraph{Ensemble Size Conclusion} In domains such as D4RL Gym where offline datasets are qualitatively similar to imitation learning datasets, larger ensembles do not result in noticeable gains. In domains such as D4RL antmaze which contain more data diversity, larger ensembles significantly improve the performance of agents.
    

\subsection{Efficient Ensembles}
    \label{sec:efficient_ensembles}
    \begin{figure*}[t]
        \centering
        \includegraphics[width=\textwidth]{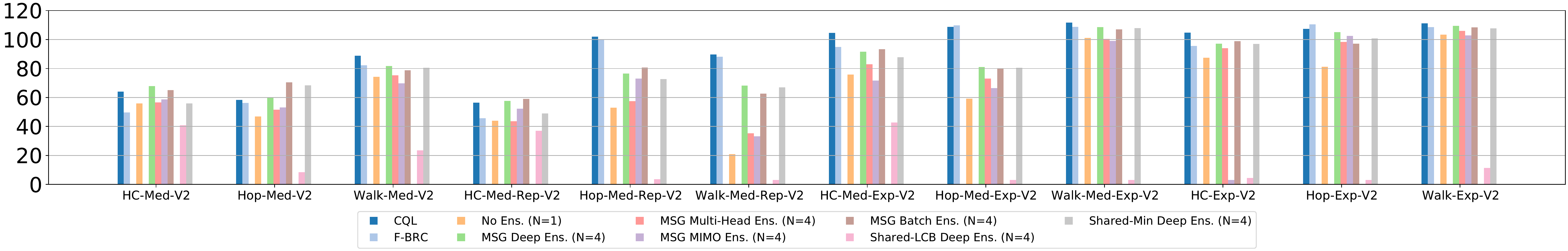}
        \includegraphics[width=\textwidth]{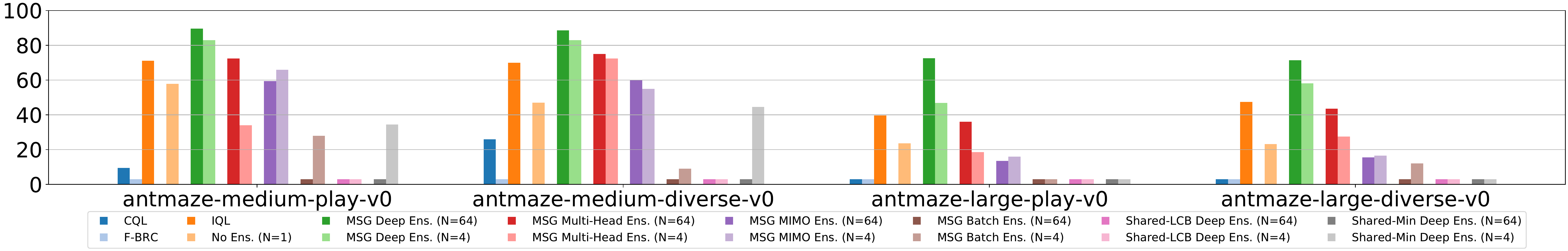}
        \caption{\small{
            Summary of D4RL benchmark results (full results presented in Appendices \ref{app:d4rl_gym} and \ref{app:d4rl_antmaze}). For each method, we report the mean across random seeds for the best hyperparameter. Numerical results for all experiments are available in our open-sourced code repository.
        }}
        \label{fig:summarrized_bar_plots}
    \end{figure*}

    Thus far we have demonstrated the significant performance gains attainable through MSG. An important concern however, is that of parameter and computational efficiency: Deep ensembles of $Q$-networks result in an $N$-fold increase in memory and compute usage, both in the policy evaluation and policy optimization phases of actor-critic training. While this might not be a significant problem in offline RL benchmark domains due to small model footprints\footnote{All our experiments were ran on a single Nvidia P100 GPU}, it becomes a major bottleneck with larger architectures such as those used in language and vision domains. To this end, we evaluate whether recent advances in ``Efficient Ensemble" approaches from the supervised learning literature transfer well to the problem of offline RL. Specifically, the efficient ensemble approaches we consider are:
    
    \paragraph{Multi-Head \citep{lee2015m, osband2016deep, tran2020hydra}} Multi-Head refers to ensembles that share a ``trunk" network and have separate ``head" networks for each ensemble member. In this work, we modify the last layer of a $Q$-network to output $N$ predictions instead of a single $Q$-value, making the computational cost of this ensemble on par with a single network.
    
    \paragraph{Multi-Input Multi-Output (MIMO) \citep{havasi2020training}} MIMO is an ensembling approach that approximately has the same parameter and computational footprint as a single network. The MIMO approach only modifies the input and output layers of a given network. In MIMO, to compute predictions for a data-point $x$ under an ensemble of size $N$, $x$ is copied $N$ times into $x_1, ..., x_N$, which are then concatenated and passed to the network. The output of the network is a vector of size $N$, which is split into $N$ predictions $y_1, ..., y_N$ representing the predictions of the ensemble members. For added clarification, we include Figure \ref{fig:mimo} depicting how a MIMO ensemble network functions. For further details on how MIMO networks are trained, we refer the interested reader to the original work of \citet{havasi2020training}.
    
    
    \paragraph{Batch Ensembles \citep{wen2020batchensemble}} Batch Ensembles incorporate rank-1 modulations to the weights of fully-connected layers. More specifically, let $W$ be the weight matrix of a given fully-connected layer, and let $x$ be the input to the layer. The output of the layer for ensemble member $i$ is computed as $\sigma(((W^T(x \circ r^i)) \circ s^i) + b^i)$, where $\circ$ is the element-wise product, parameters with superscript $i$ are separate for each ensemble member, and $\sigma$ is the activation function. While Batch Ensembles is efficient in terms of number of parameters, in our actor-critic setup its computational cost is similar to deep ensembles, since for policy updates we need to evaluate every ensemble member using separate forward passes.

    A runtime comparison of different ensembling approaches can be viewed in Table \ref{tab:runtimes}.
    
    \paragraph{D4RL Gym Domains}
    Appendix \ref{app:gym_efficient_ensembles} presents our results in the D4RL Gym domains with ensemble size $N=4$ (summary in Figure \ref{fig:summarrized_bar_plots}). Amongst the considered efficient ensemble approaches, Batch Ensembles~\citep{wen2020batchensemble} result in the best performance, which follows findings from the supervised learning literature~\citep{ovadia2019can}.
    
    \paragraph{D4RL Antmaze Domains} Appendix \ref{app:antmaze_efficient_ensembles} presents our results in the D4RL antmaze domains for both ensemble sizes of $N=4$ and $N=64$ (summary in Figure \ref{fig:summarrized_bar_plots}). As can be seen, compared to MSG with deep ensembles (separate networks), the efficient ensemble approaches we consider are very unreliable, and fail for most hyperparameter choices.
    
    
    
    
    \paragraph{Efficient Ensembles Conclusion}
    We believe the observations in this section very clearly motivate future work in developing efficient uncertainty estimation approaches that are better suited to the domain of reinforcement learning. To facilitate this direction of research, in our open-sourced codebase we have included a complete boilerplate example of an offline RL agent, amenable to drop-in implementation of novel uncertainty-estimation techniques.

    \begin{table}[t]
    \centering
    \small{
        \begin{tabular}{|l|c|}
            \hline
            \textbf{Method}              & \textbf{Runtime}                                        \\ \hline
            No Ens. ($N=1$) ($\alpha > 0$)& $1\times$                                               \\ \hline
            CQL                          & \begin{tabular}[c]{@{}c@{}}$0.27\times$\end{tabular} \\ \hline
            F-BRC                        & \begin{tabular}[c]{@{}c@{}}$0.22\times$\end{tabular} \\ \hline
            MSG Deep Ens. ($N=4$)        & \begin{tabular}[c]{@{}c@{}}$0.64\times$\end{tabular} \\
            MSG Multi-Head Ens. ($N=4$)  & \begin{tabular}[c]{@{}c@{}}$0.92\times$\end{tabular} \\
            MSG MIMO Ens. ($N=4$)        & \begin{tabular}[c]{@{}c@{}}$0.90\times$\end{tabular} \\
            MSG Batch Ens. ($N=4$)       & \begin{tabular}[c]{@{}c@{}}$0.72\times$\end{tabular} \\ \hline
            MSG Deep Ens. ($N=64$)       & \begin{tabular}[c]{@{}c@{}}$0.11\times$\end{tabular} \\
            MSG Multi-Head Ens. ($N=64$) & \begin{tabular}[c]{@{}c@{}}$0.88\times$\end{tabular} \\
            MSG MIMO Ens. ($N=64$)       & \begin{tabular}[c]{@{}c@{}}$0.61\times$\end{tabular} \\
            MSG Batch Ens. ($N=64$)      & \begin{tabular}[c]{@{}c@{}}$0.29\times$\end{tabular} \\ \hline
        \end{tabular}
    }
    \caption{\small{
        Runtime comparison of various approaches. We report training iterations per second, relative to no ensembling while using the CQL-style regularizer described in Section \ref{sec:tradeoff}. All methods were trained on an Nvidia P100 GPU.
    }}
    \label{tab:runtimes}
\end{table}

    
    


    
    


\section{Practical Workflow for Applying MSG to New Offline RL Datasets}    
    
    
    Throughout this work, we have reported the results for every method, domain, hyperparameter choice, and random seed. Based on our experience, in this section we outline a practical workflow for applying MSG to new offline RL environments and datasets.
    
    In addition to the the key take-aways from our discussion of Figure \ref{fig:ens_size_no_sass} in Section \ref{sec:ablations},
    here we include practical advice on hyperparameter tuning when applying MSG to a new domain. Aside from the choice of neural network architectures, the main hyperparameters in MSG are $\beta$, i.e. the hyperparameter controlling the amount of pessimism in $Q_{LCB}$ (Equation \ref{eq:msg_policy_objective}), and $\alpha$, i.e. the hyperparameter controlling the contribution of the CQL-style regularizer (Equation \ref{eq:full_msg_q_objective}).
    Intuitively, as described in Section \ref{sec:tradeoff}, larger values of $\alpha$ become necessary when the offline dataset has a narrow distribution (e.g. imitation learning datasets), or in MDPs with more chaotic dynamics, where small deviations can be very costly and thus we must stay close to the provided data support.
    Equipped with this intuition, for a new offline RL task and dataset, we would first use low $\beta$ values (e.g. $\beta \in \{-4, -8\}$ or lower), and search for an appropriate range of $\alpha$ (which may be $0$, as in our reported antmaze-large results in Table \ref{tab:hard_tasks}). For further hyperparameter tuning from this starting point, we would investigate if reducing pessimism by increasing $\beta$ -- and potentially modifying $\alpha$ -- would lead to improved policies.
    Generally, we find MSG to be fairly robust, as evidenced by our results in Appendices \ref{app:d4rl_gym} and \ref{app:d4rl_antmaze}, and Figure \ref{fig:ens_size_no_sass}.
    
    

\section{Discussion \& Future Work}
Our work has highlighted the significant power of ensembling as a mechanism for uncertainty estimation for offline RL. In this work we took a renewed look into $Q$-ensembles, and studied how to leverage them as the primary source of pessimism for offline RL.
Through theoretical analyses and toy constructions, we demonstrated a critical flaw in the popular approach of using shared targets for obtaining pessimistic $Q$-values, and demonstrated that it can in fact lead to optimistic estimates. Using a simple fix, we developed a practical deep offline RL algorithm, MSG, which resulted in large performance gains on established offline RL benchmarks.

As demonstrated by our experimental results, an important outstanding direction is to study how we can design improved efficient ensemble approximations, as we have demonstrated that current approaches used in supervised learning
are not nearly as effective as MSG with ensembles that use separate networks. We hope that this work engenders new efforts from the community of neural network uncertainty estimation researchers towards developing efficient uncertainty estimation techniques directed at reinforcement learning. 



Lastly, inspired by the EMaQ algorithm~\citep{ghasemipour2021emaq}, an intriguing opportunity for future work is to leverage MSG as the ``backend" learning algorithm in online RL\footnote{Another important setting is the scenario of finetuning offline RL trained agents using online RL.}, where actors collect new data throughout training. In online RL effective exploration is critical, and uncertainty estimation has a long history of being used to devise exploration heuristics~\citep{ghavamzadeh2016bayesian}. Similar to the works of \citet{chen2017ucb} and \citet{osband2016deep}, the ensemble of independently trained $Q$-functions in MSG can be used to compute an Upper Confidence Bound (UCB) criterion for choosing exploratory actions, which may lead to a strong exploration technique that leverages the inductive biases of the neural network architectures used.

    

\subsubsection*{Acknowledgments}
We would like to thank Yasaman Bahri for insightful discussions regarding infinite-width neural networks. We would like to thank Laura Graesser for providing a detailed review of our work. We would like to thank conference reviewers for posing important questions that helped clarify the organization of this manuscript.




\bibliography{iclr2022_conference}
\bibliographystyle{icml2022}

\newpage
\appendix
\onecolumn
\clearpage

\section{D4RL Gym Locomotion Benchmarks}
\label{app:d4rl_gym}
    In this section we discuss results using the D4RL gym (\texttt{halfcheetah}, \texttt{hopper}, \texttt{walker2d}) benchmark domains. The dataset types that we consider are \texttt{medium-v2}, \texttt{medium-replay-v2}, \texttt{medium-expert-v2}, and \texttt{expert-v2}.
    
    \subsection{Experimental Details}
        \label{app:gym_details}
        All policies and Q-functions are a 3 layer neural network with relu activations and hidden layer size 256. The policy output is a normal distribution that is squashed to $[-1, 1]$ using the tanh function. All methods were trained for 3M steps. CQL and MSG are trained with behavioral cloning (BC) for the first 50K steps. F-BRC pretrains a behavioral cloning model for 1M steps.
        
        MSG, CQL, and F-BRC, are tuned with an equal hyperparameter search budget of 12 hyperparameter choices. Each hyperparameter choice for each method is trained using 2 random seeds. Each run is evaluated for 100 episodes. We report results for all hyperparameters and all seeds.
        For fairness of comparison, F-BRC is ran without adding a survival reward bonus. MSG and CQL are implemented in our code, and for F-BRC we use the opensourced codebase. Our reported CQL results appear to be better than or on par with values reported in prior works, which provided us with confidence to use our own implementation. We used the following values for hyperparameter search:
        
        \paragraph{MSG} $\beta \in \{-4. -8.\}$, $\alpha \in \{0., 0.5, 1., 2., 4., 8.\}$
        \paragraph{CQL} $\alpha \in \{0., 0.45, 0.9, 1.36, 1.81, 2.27, 2.32, 3.18, 3.63, 4.09, 4.55, 5.\}$
        \paragraph{F-BRC} $\lambda \in \{0., 0.1, 0.2, 0.3, 0.4, 0.5, 0.6, 0.7, 0.8, 0.9, 1.0, 1.1\}$

    \subsection{Baseline Comparison}
        \label{app:gym_baselines}
\begin{figure}[t]
            \centering
            \includegraphics[width=\linewidth]{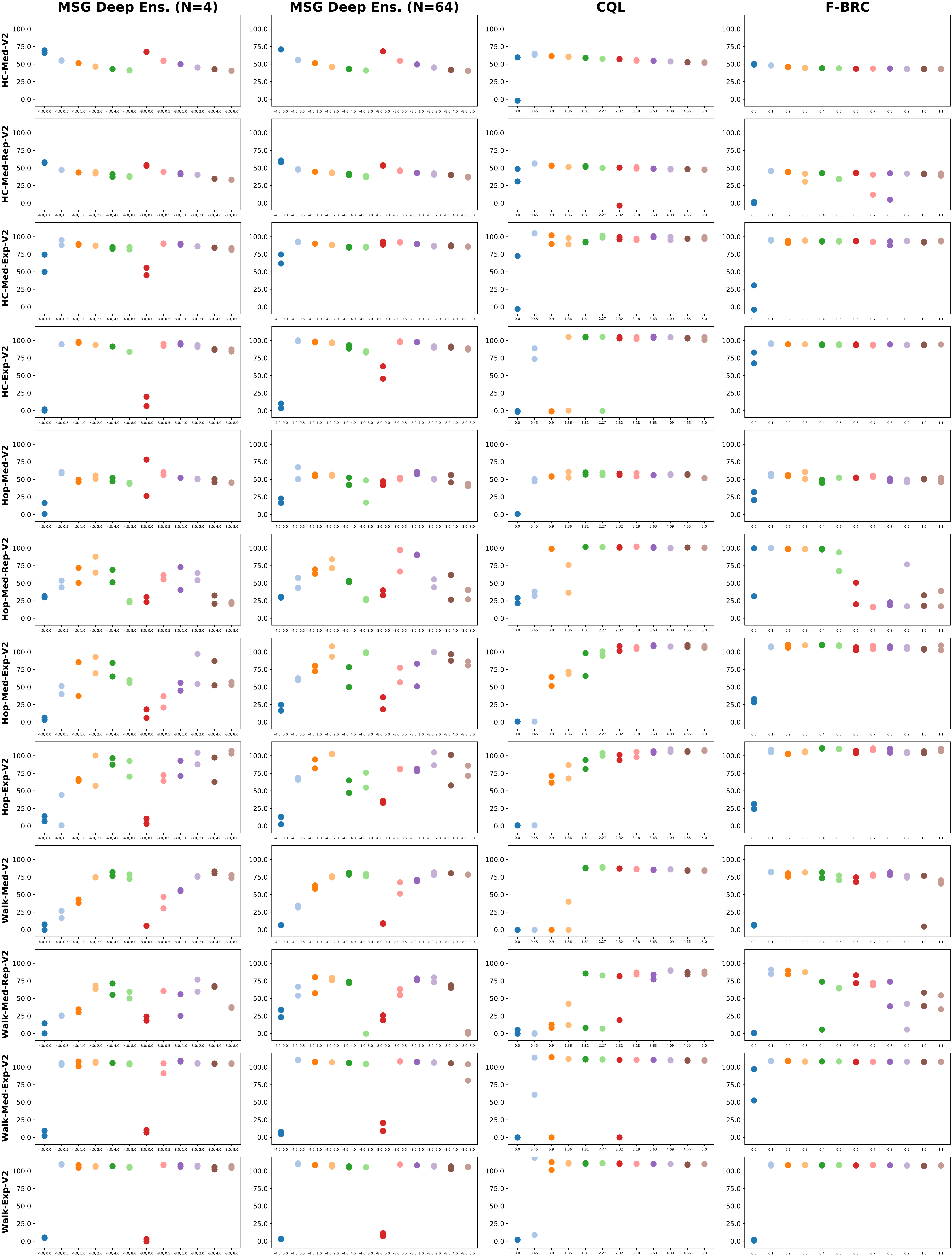}
            \caption{Basline comparison on D4RL Gym domains. The hyperparameters for MSG, CQL, and F-BRC are $(\beta, \alpha)$, $\alpha$, and $\lambda$ respectively. For legibility of hyperparameter values on the $x$-axis, please zoom into the pdf document.}
            \label{fig:gym_baselines}
        \end{figure}

        We compare MSG with deep ensembles to CQL~\citep{kumar2020conservative} and F-BRC~\citep{kostrikov2021offline}, two methods that obtain state-of-the-art results on the gym domains of the D4RL benchmark. Figure \ref{fig:gym_baselines} presents our results on the gym domains. The key takeaways of our results are as follows:
        \begin{itemize}
            \item On gym domains, MSG is generally competitive with current state-of-the-art results.
            \item In the \texttt{hopper} and \texttt{walker2d} domains -- where dynamics can be chaotic and staying close to the data-support is beneficial -- CQL and F-BRC which rely on support constraints are more reliable w.r.t. the hyperparameter choice.
        \end{itemize}
    
    \subsection{Ablations}
        \label{app:gym_ablations}
\begin{figure}[t]
            \centering
            \includegraphics[width=\linewidth]{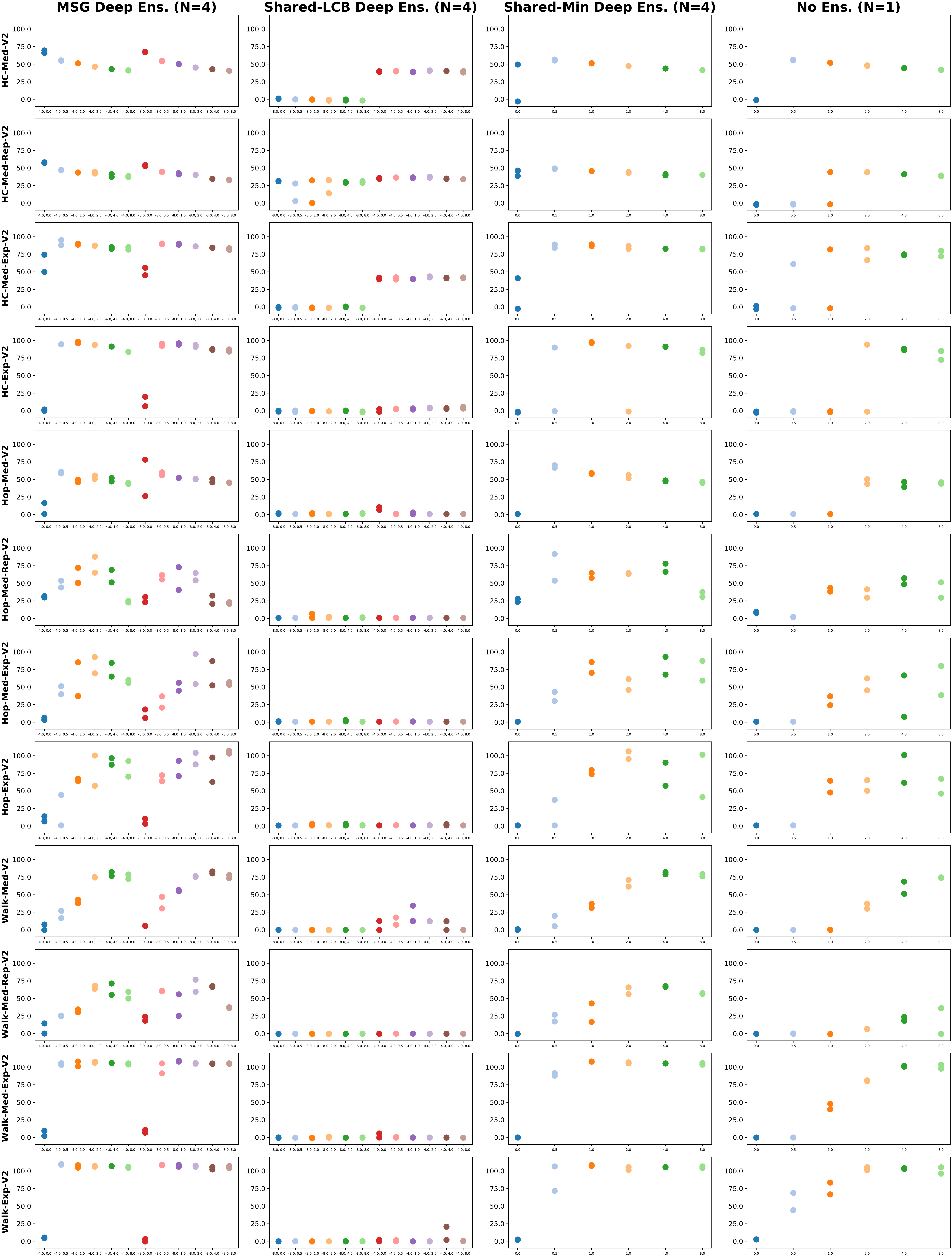}
            \caption{Ablations on D4RL Gym domains. The hyperparameters for MSG, Shared-LCB, Shared-Min, and No Ensemble are $(\beta, \alpha)$, $(\beta, \alpha)$, $\alpha$, $\alpha$ respectively. For legibility of hyperparameter values on the $x$-axis, please zoom into the pdf document.}
            \label{fig:gym_ablations}
        \end{figure}

\begin{figure}[t]
            \centering
            \includegraphics[width=\linewidth]{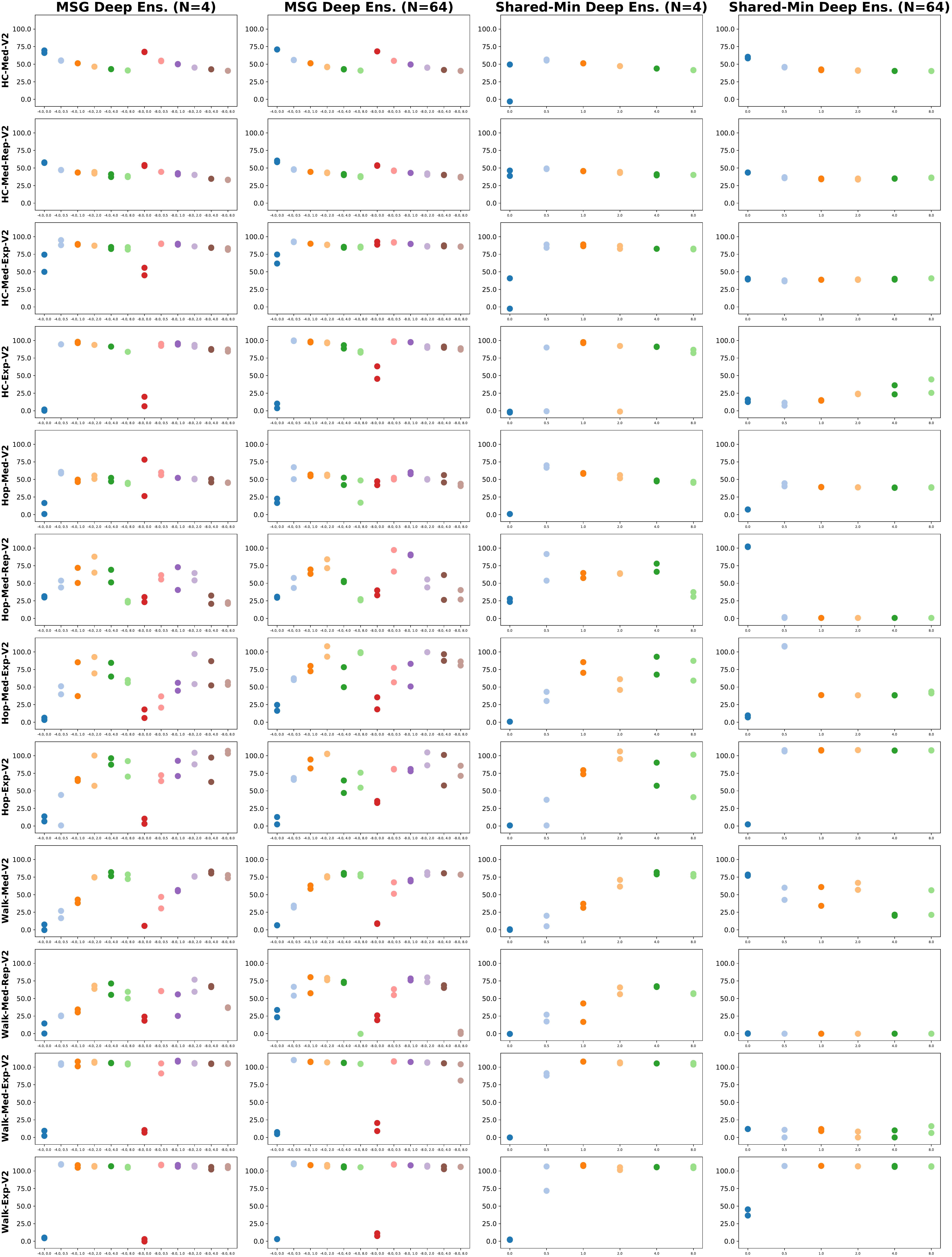}
            \caption{Comparison of MSG and Shared-Min on D4RL Gym domains, for two ensemble sizes of $N=4$ and $N=64$. The hyperparameters for MSG and Shared-Min are $(\beta, \alpha)$ and $\alpha$ respectively. For legibility of hyperparameter values on the $x$-axis, please zoom into the pdf document.}
            \label{fig:gym_msg_vs_shared_min}
        \end{figure}

        We perform ablations w.r.t. the key components of MSG. Our key takeaways from the results in Figures \ref{fig:gym_ablations} and \ref{fig:gym_msg_vs_shared_min} for gym domains are as follows:
        \begin{itemize}
            \item Comparing MSG -- which uses independent targets -- to Shared-LCB and Shared-Min -- which use shared targets -- clearly demonstrates the significance of our theoretical analysis in Section \ref{sec:independence_matters}. Despite differing in only 2 lines of code from MSG, Shared-LCB and Shared-Min very significantly underpeform MSG. In fact, using no ensembling at all $(N=1)$ outperforms Shared-LCB and is on par with Shared-Min.
            \item In the gym domains of D4RL -- which are close to being imitation learning datasets -- using large ensemble sizes does not result in noticeable gains in MSG.
        \end{itemize}
    
    \subsection{Efficient Ensembles}
        \label{app:gym_efficient_ensembles}
\begin{figure}[t]
            \centering
            \includegraphics[width=\linewidth]{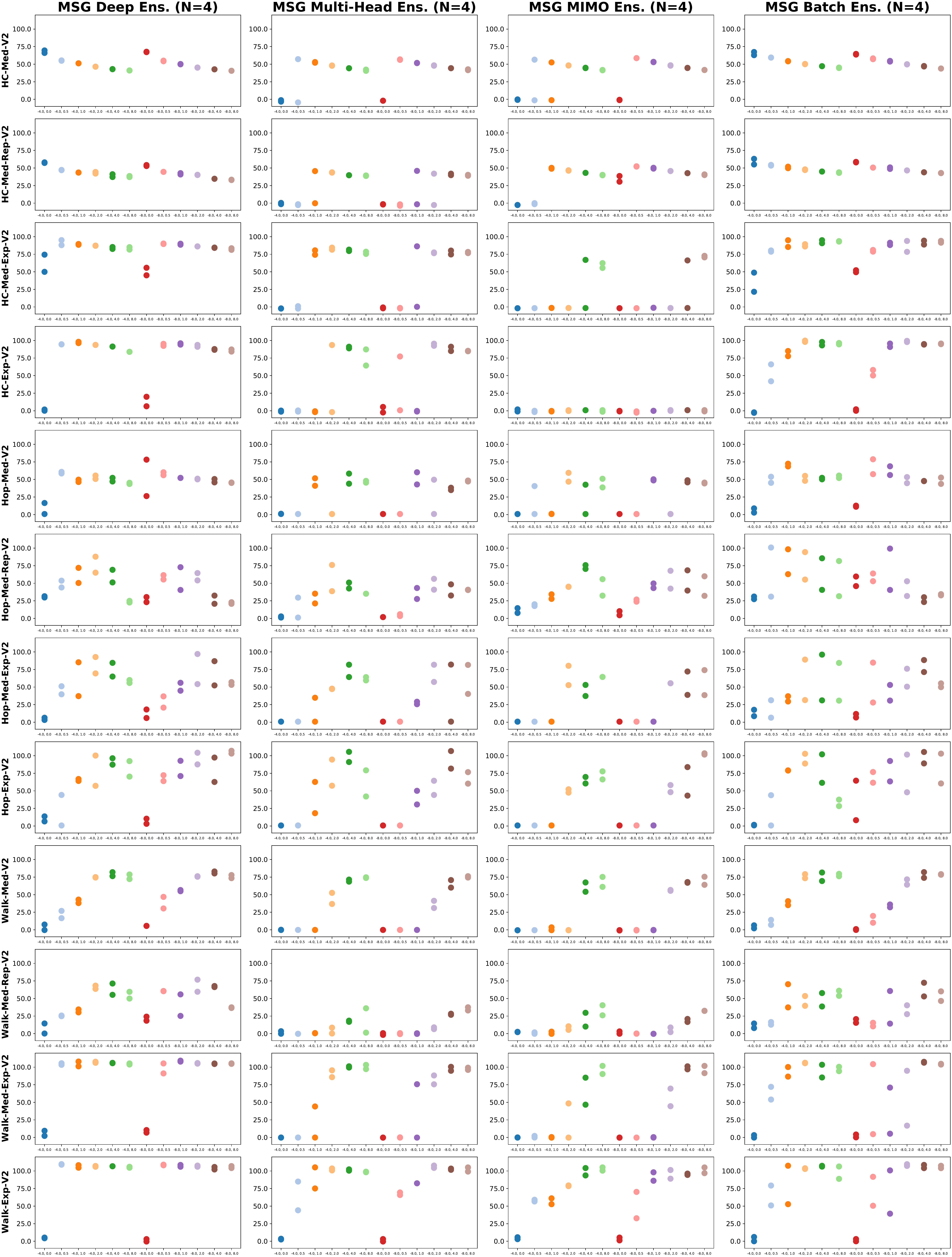}
            \caption{Efficient ensembles on D4RL Gym domains. The hyperparameters for MSG are $(\beta, \alpha)$. For legibility of hyperparameter values on the $x$-axis, please zoom into the pdf document.}
            \label{fig:gym_efficient_ensembles}
        \end{figure}

        As discussed in Section \ref{sec:efficient_ensembles}, we evaluate whether the performance of MSG with deep ensembles can be matched using state-of-the-art ``efficient ensembles" from supervised learning literature. Our takeaways from the results in Figure \ref{fig:gym_efficient_ensembles} for gym domains are as follows:
        \begin{itemize}
            \item From the efficient ensembling approaches considered, Batch Ensembles tend to be the most performant \citep{wen2020batchensemble}. Interestingly, this follows the findings of \citet{ovadia2019can} from the supervised learning literature.
        \end{itemize}

\section{D4RL Antmaze Benchmarks}
\label{app:d4rl_antmaze}
    In this section we discuss results using the D4RL antmaze benchmark domains. We mainly focus our experiments on the \texttt{antmaze-medium} and \texttt{antmaze-large} environments, with the two available offline datasets of \texttt{play-v0} and \texttt{diverse-v0}.
    
    \subsection{Experimental Details}
        The rewards in the antmaze domains (which are either $0.$ or $1.$) were transformed using the following equation: $4(r - 0.5)$. This transformation has become common practice in prior works \url{https://github.com/aviralkumar2907/CQL/blob/d67dbe9cf5d2b96e3b462b6146f249b3d6569796/d4rl/examples/cql_antmaze_new.py#L22}.
        
        All policies and Q-functions are a 3 layer neural network with relu activations and hidden layer size 256. The policy output is a normal distribution that is squashed to $[-1, 1]$ using the tanh function. All methods were trained for 2M steps. CQL and MSG are trained with behavioral cloning (BC) for the first 50K steps. F-BRC pretrains a behavioral cloning model for 1M steps.
        
        MSG, CQL, and F-BRC, are tuned with an equal hyperparameter search budget of 8 hyperparameter choices. Each run is evaluated for 100 episodes. We report results for all hyperparameters and all seeds.
        
        For all MSG with deep ensembles experiments (that appear in Figure \ref{fig:ens_size_no_sass} also), and for the No Ensembles ($N=1$) baseline we ran 5 random seeds per hyperparameter choice. For all other baselines and ablations that appear in this section we ran 2 random seeds.
        
        For fairness of comparison, F-BRC is ran without adding a survival reward bonus. MSG and CQL are implemented in our code, and for F-BRC we use the opensourced codebase. Our reported CQL results appear to be better than or on par with values reported in prior works, which provided us with confidence to use our own implementation. We used the following values for hyperparameter search:
        
        \paragraph{MSG} $\beta \in \{-4. -8.\}$, $\alpha \in \{0., 0.1, 0.5, 1.\}$
        \paragraph{CQL} $\alpha \in \{0., 0.45, 0.9, 1.36, 1.81, 2.27, 2.32, 3.18, 3.63, 4.09, 4.55, 5.\}$
        \paragraph{F-BRC} $\lambda \in \{0., 0.14, 0.29, 0.43, 0.57, 0.71, 0.86, 1.\}$

    \subsection{Baseline Comparison}
        \label{app:antmaze_baselines}
\begin{figure}[H]
            \centering
            \includegraphics[width=\linewidth]{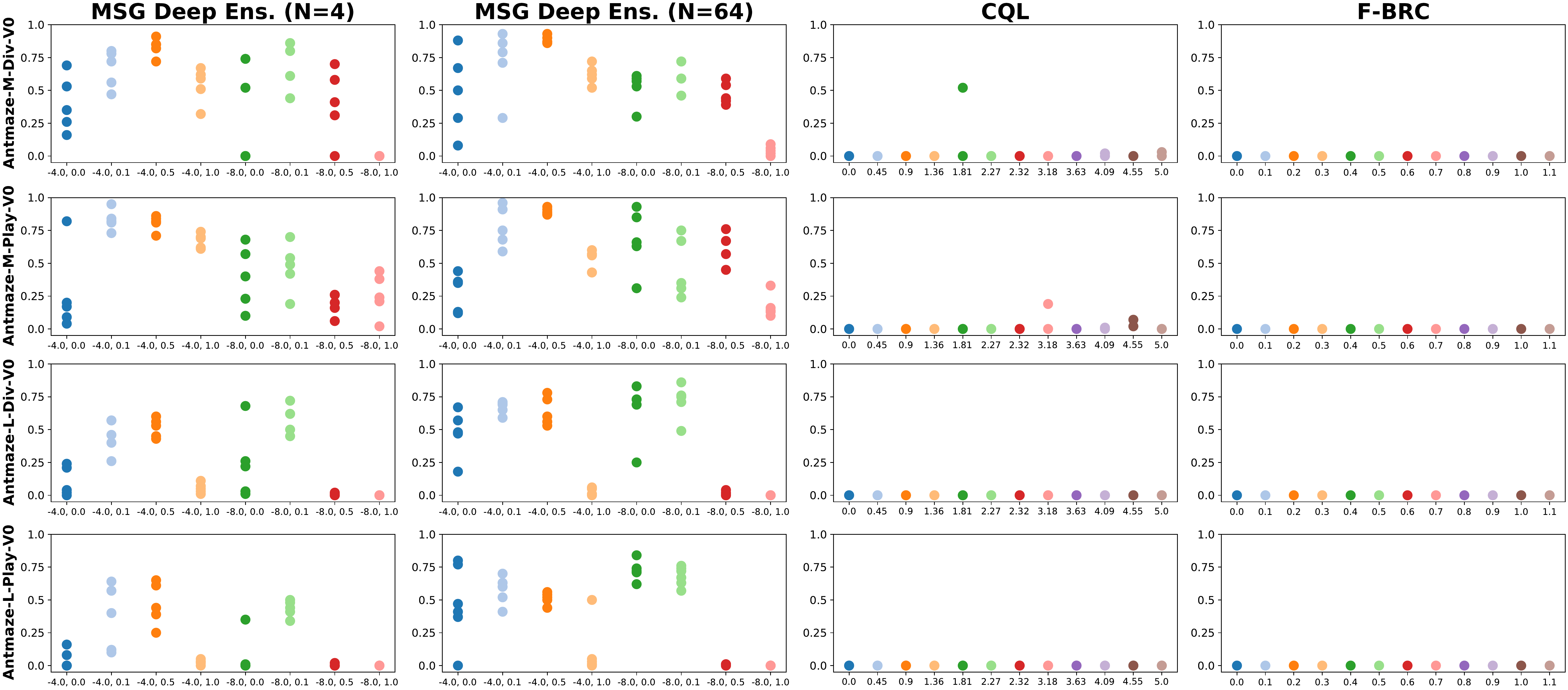}
            \caption{Basline comparison on D4RL antmaze domains. The hyperparameters for MSG, CQL, and F-BRC are $(\beta, \alpha)$, $\alpha$, and $\lambda$ respectively. For legibility of hyperparameter values on the $x$-axis, please zoom into the pdf document.}
            \label{fig:antmaze_baselines}
        \end{figure}

        We compare MSG with deep ensembles to CQL \citep{kumar2020conservative} and F-BRC\citep{kostrikov2021offline}. Figure \ref{fig:antmaze_baselines} presents our results on the antmaze domains. The key takeaways of our results are as follows:
        \begin{itemize}
            \item On the D4RL antmaze domains, MSG with deep ensembles far exceeds the results of prior state of the art result Implicit Q-Learning (IQL) \citep{kostrikov2021offlineIQL}.
            \item CQL \citep{kumar2020conservative} and F-BRC \citep{kostrikov2021offline} which obtain very strong results on the D4RL Gym domains (Appendix \ref{app:gym_baselines}), completely fail on the D4RL antmaze domains\footnote{Note that our implementation of CQL exceeds the reported results of the original paper on D4RL Gym domains, providing us with confidence in our implementation. For D4RL antmaze domains we have also experimented with the open-sourced codebase accompanying \citet{kostrikov2021offline}, and the open-sourced codebase accompanying the original CQL work \citep{kumar2020conservative}, but were unable to obtain the results reported in original paper.} This mimics the results of \citet{kostrikov2021offlineIQL} that demonstrated many prior offline RL methods that perform strongly on D4RL Gym domains, fail on D4RL antmaze domains. We believe this highlights the importance of using more complex offline RL benchmarks that emphasize the stiching of diverse trajectories through dynamic programming, as opposed to offline RL datasets for Gym domains that are qualitatively very similar to imitation learning datasets.
        \end{itemize}
    
    \subsection{Ablations}
        \label{app:antmaze_ablations}
\begin{figure}[H]
            \centering
            \includegraphics[width=\linewidth]{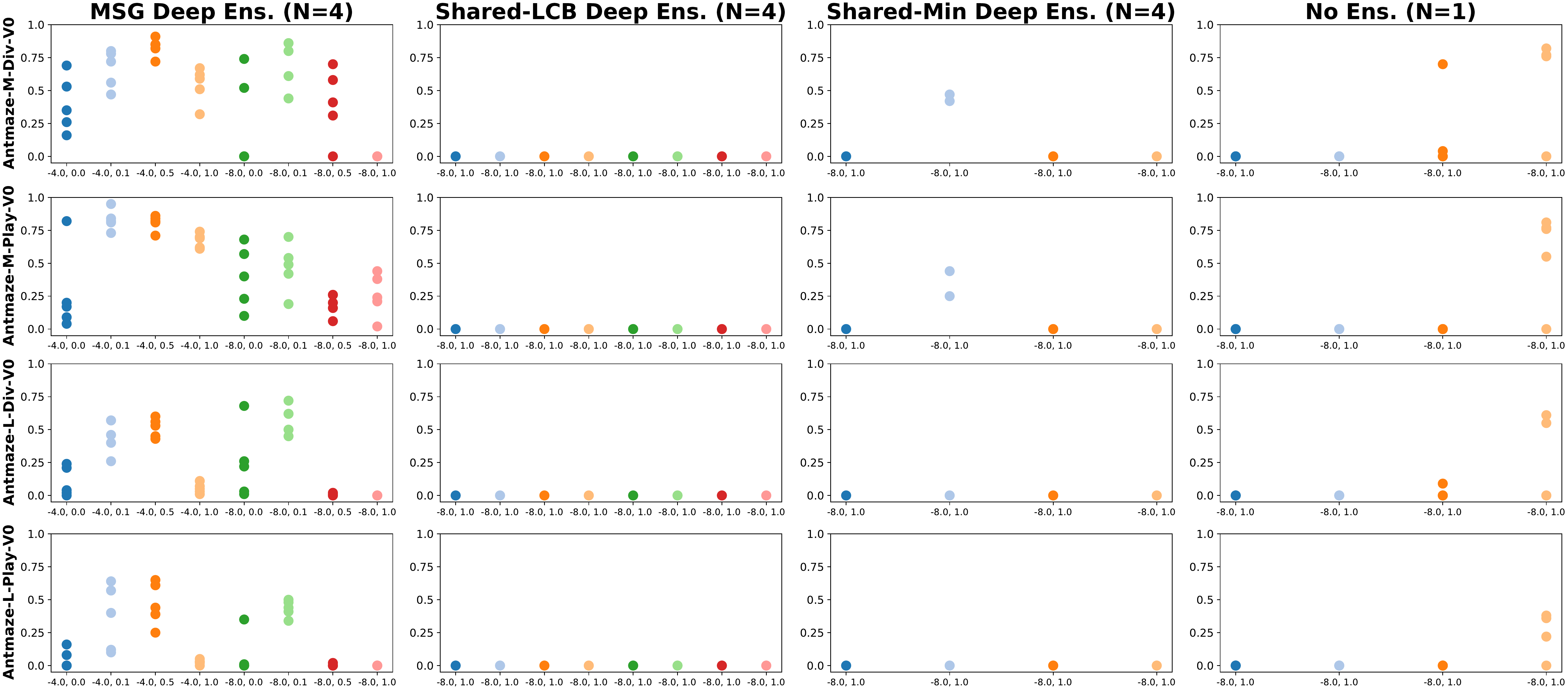}
            \caption{Ablations on D4RL antmaze domains ($N=4$). The hyperparameters for MSG, Shared-LCB, Shared-Min, and No Ensemble are $(\beta, \alpha)$, $(\beta, \alpha)$, $\alpha$, $\alpha$ respectively. For legibility of hyperparameter values on the $x$-axis, please zoom into the pdf document.}
            \label{fig:antmaze_ablations_N_4}
        \end{figure}

\begin{figure}[H]
            \centering
            \includegraphics[width=\linewidth]{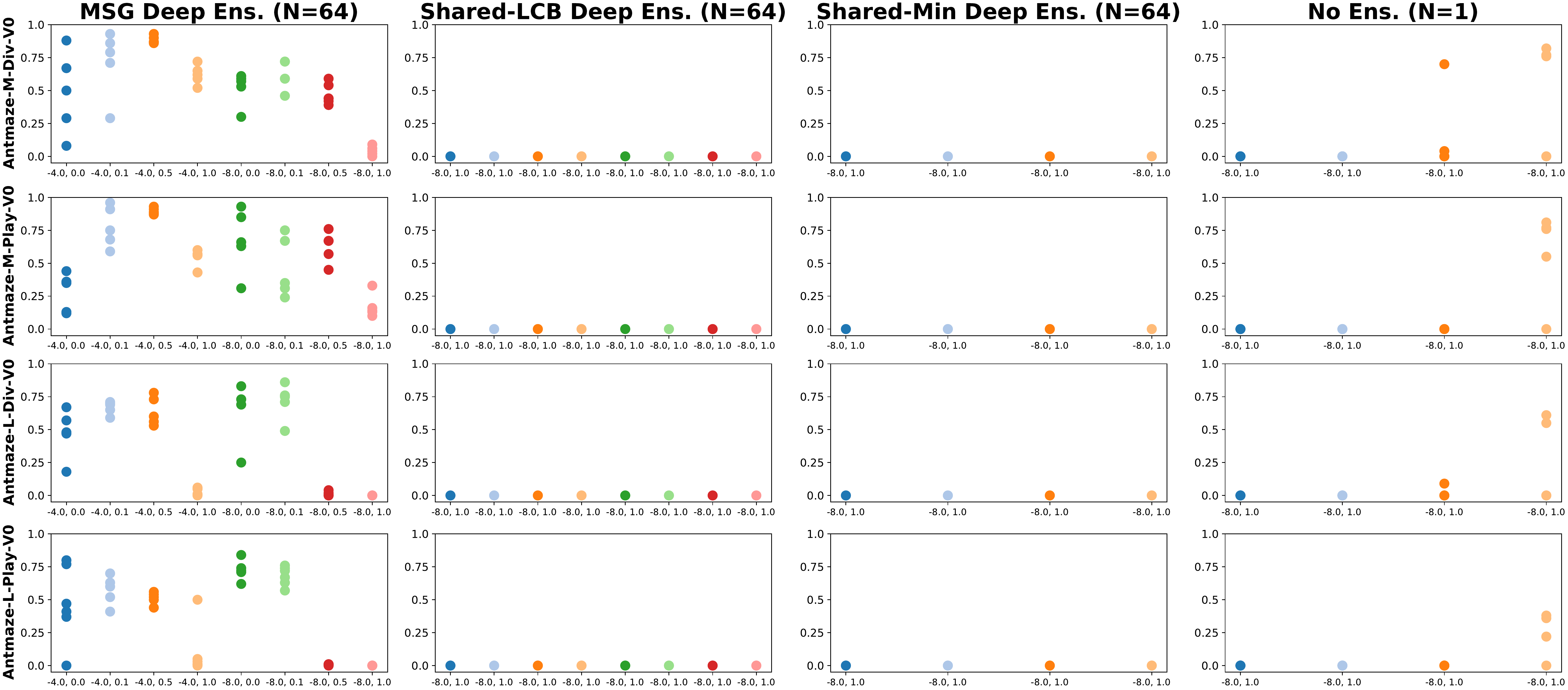}
            \caption{Ablations on D4RL antmaze domains ($N=64$). The hyperparameters for MSG, Shared-LCB, Shared-Min, and No Ensemble are $(\beta, \alpha)$, $(\beta, \alpha)$, $\alpha$, $\alpha$ respectively. For legibility of hyperparameter values on the $x$-axis, please zoom into the pdf document.}
            \label{fig:antmaze_ablations_N_64}
        \end{figure}

        We perform ablations w.r.t. the key components of MSG. Our key takeaways from the results in Figures \ref{fig:antmaze_ablations_N_4} and \ref{fig:antmaze_ablations_N_64} for antmaze domains are as follows:
        \begin{itemize}
            \item Comparing MSG -- which uses independent targets -- to Shared-LCB and Shared-Min -- which use shared targets -- clearly demonstrates the significance of our theoretical analysis in Section \ref{sec:independence_matters}. Despite differing in only 2 lines of code from MSG, Shared-LCB and Shared-Min very significantly underpeform MSG. In fact, using no ensembling at all $(N=1)$ outperforms Shared-LCB and is on par with Shared-Min.
            \item In the gym domains of D4RL -- which are close to being imitation learning datasets -- using large ensemble sizes does not result in noticeable gains in MSG.
        \end{itemize}
    
    \subsection{Efficient Ensembles}
        \label{app:antmaze_efficient_ensembles}
\begin{figure}[H]
            \centering
            \includegraphics[width=\linewidth]{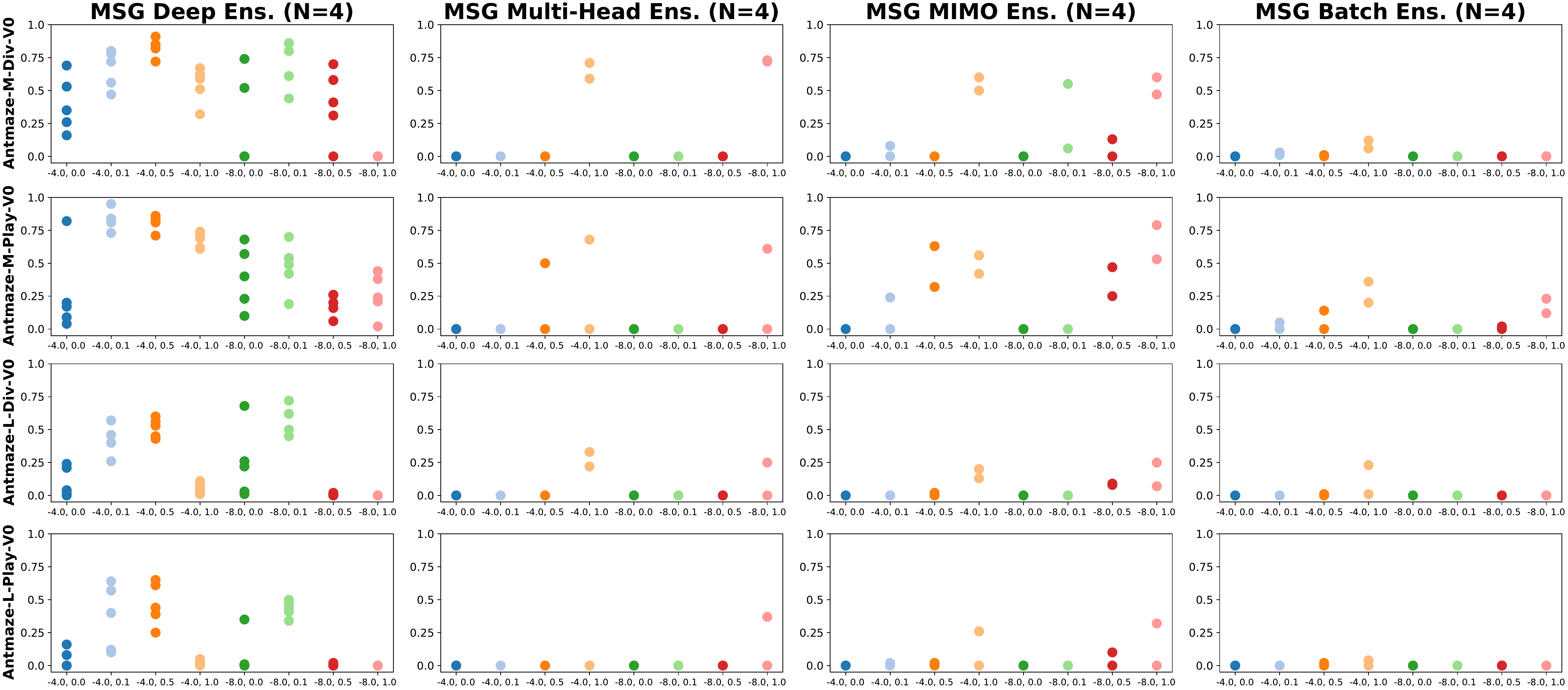}
            \caption{Efficient ensembles on D4RL antmaze domains ($N=4$). The hyperparameters for MSG are $(\beta, \alpha)$. For legibility of hyperparameter values on the $x$-axis, please zoom into the pdf document.}
            \label{fig:antmaze_efficient_ensembles_N_4}
        \end{figure}

\begin{figure}[H]
            \centering
            \includegraphics[width=\linewidth]{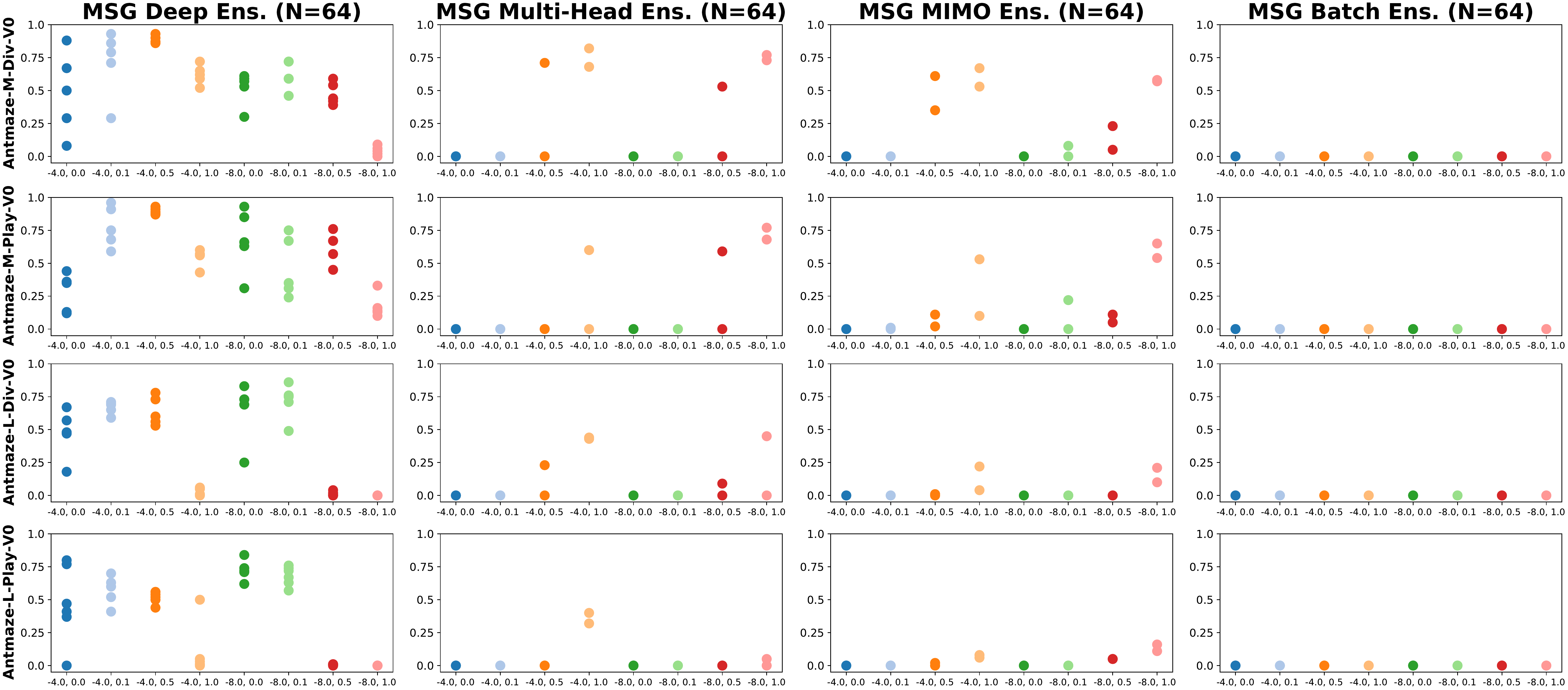}
            \caption{Efficient ensembles on D4RL antmaze domains ($N=64$). The hyperparameters for MSG are $(\beta, \alpha)$. For legibility of hyperparameter values on the $x$-axis, please zoom into the pdf document.}
            \label{fig:antmaze_efficient_ensembles_N_64}
        \end{figure}

        As discussed in Section \ref{sec:efficient_ensembles}, we evaluate whether the performance of MSG with deep ensembles can be matched using state-of-the-art ``efficient ensembles" from supervised learning literature. Our takeaways from the results in Figures \ref{fig:antmaze_efficient_ensembles_N_4} and \ref{fig:antmaze_efficient_ensembles_N_64} for antmaze domains are as follows:
        \begin{itemize}
            \item Despite efficient ensembles such as Batch Ensembles performing well on the D4RL Gym domains (Appendix \ref{app:gym_efficient_ensembles}), they fail on the antmaze domains for both ensemble sizes of $N=4$ and $N=64$.
        \end{itemize}

\section{RL Unplugged}
    \label{app:rlu}
    
    
    \subsection{DM Control Suite Tasks}
        The networks used in \citet{gulcehre2020rl} for DM Control Suite Tasks are very large relative to the networks we used in the D4RL benchmark; roughly the networks contain 60x more parameters. Using a large ensemble size with such architectures requires training using a large number of devices. Furthermore, since in our experiments with efficient ensemble approximations we did not find a suitable alternative to deep ensembles (section \ref{sec:efficient_ensembles}), we decided to use the same network architectures and $N = 64$ as in the D4RL setting (enabling single-GPU training as before).
        
        Our hyperparameter search procedure was similar to before, where we first performed a coarse search using 2 random seeds and hyperparameters $\beta \in \{-1., -2., -4. -8.\}$, $\alpha \in \{0., 0.5\}$, and for the best found hyperparameter, ran final experiments with 5 new random seeds.

\section{Algorithm Box}
    \begin{algorithm2e}[H]
      \DontPrintSemicolon
      
      \textbf{Input:} offline RL dataset $D$, $Q$-function and policy architectures, ensemble size $N$, MSG hyperparameters $\beta, \alpha$\;
      
      Create $N$ $Q$-networks $Q_{\theta^i}$, with parameters sampled from the initial weight distribution\;
      Create target networks $Q_{\bar{\theta^i}}$, initialized as $\bar{\theta^i} \leftarrow \theta^i$\;
      Create the policy network $\pi$ with parameters $\theta^\pi$\;
      
      \SetKwProg{Fn}{Function}{:}{}
      
      \SetKwFunction{fPE}{PolicyEvaluationStep}
        \Fn{\fPE{batch}}{
            \For{$i=1,...,N$}{
                \tcc{Compute policy evaluation loss}
                $\forall s'_m \in batch, a'_{\pi, m} \sim \pi(s'_m)$\;
                $L(\theta_i) = \frac{1}{M} \sum_m \Big(Q_{\theta^i}(s_m,a_m) - (r + \gamma \cdot Q_{\bar{\theta^i}}(s'_m, a'_{\pi, m}))\Big)^2$\;
                \tcc{Compute support constraint regularizer loss}
                $\forall s_m \in batch, a_{\pi, m} \sim \pi(s'_m)$\;
                $\mathcal{H}(\theta_i) = \frac{1}{M} \sum_m Q_{\theta^i}(s_m,a_{\pi, m}) - Q_{\theta^i}(s_m,a_m)$\;
                \tcc{Optimize the policy evaluation objective}
                $\mathcal{L}(\theta^i) = L(\theta^i) + \alpha \cdot \mathcal{H}(\theta^i)$\;
                $\theta^i \leftarrow \theta^i - \texttt{AdamUpdate}\Big(\mathcal{L}(\theta^i), \theta^i\Big)$\;
                $\bar{\theta^i} \leftarrow \tau \cdot \bar{\theta^i} + (1 - \tau) \cdot \theta_i$\;
            }
        }
    
      \SetKwFunction{fPO}{MSGPolicyOptimizationStep}
      \Fn{\fPO{batch}}{
            \tcc{Compute LCB $Q$-values}
            $\forall s_m \in batch, a_{\pi, m} \sim \pi(s_m)$\;
            $\forall s_m \in batch, Q_{\mathrm{LCB}}(s_m,a_{\pi,m}) = \underset{i}{\mathrm{mean}}\left[Q_{\theta^i}(s_m,a_{\pi,m})\right] - \beta \cdot \underset{i}{\mathrm{std}}\left[Q_{\theta^i}(s_m,a_{\pi,m})\right]$\;
            \tcc{Optimize the policy objective}
            $\mathcal{L}(\theta^\pi) = \frac{1}{M} \sum_m Q_{\mathrm{LCB}}(s_m,a_{\pi,m})$\;
            $\theta^\pi \leftarrow \theta^\pi - \texttt{AdamUpdate}\Big(\mathcal{L}(\theta^\pi), \theta^\pi\Big)$\;
      }
      \;
      \SetKwFunction{fMain}{Main}
      \Fn{\fMain{}}{
            \tcc{Optional initialization of the policy using behavioral cloning while training $Q$-functions}
            \For{\text{desired number of steps}}{
                batch $\sim D$\;
                \texttt{PolicyEvaluationStep}(batch)\;
                Update $\pi$ using behavioral cloning\;
            }
            \For{\text{desired number of steps}}{
                batch $\sim D$\;
                \texttt{PolicyEvaluationStep}(batch)\;
                \texttt{MSGPolicyOptimizationStep}(batch)\;
            }
      }
      \caption{
        Pseudocode for MSG Algorithm Using Deep Ensembles
      }
      \label{alg:full_alg}
    \end{algorithm2e}

\begin{figure}[t]
    \centering
    \includegraphics[width=0.6\textwidth]{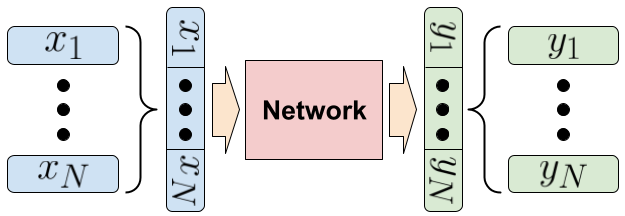}
    \caption{Visual depiction of MIMO Ensemble}
    \label{fig:mimo}
\end{figure}



\section{Theorem Proof}
    \label{app:proof}
    In this section, following our notation and inifite-width networks setup described in Section \ref{sec:math_char}, we present the proof for the following Theorem.
    
    \begin{theorem}
        For a given $(s,a) \in \mathcal{S} \times \mathcal{A}$, let $Q^{(0)}_{\theta^i}(s,a)$ denote $Q_{\theta^i}(s,a)\vert_{t=0}$ (value at initialization), with $\theta$ sampled from the initial weight distribution.
        After $t+1$ iterations of pessimistic policy evaluation, the LCB value estimate for $(s',\pi(s')) \in \exprime$ is given by,
            \begin{align}
                &\textbf{Independent Targets (Method 1):}\nonumber\\
                &Q_{\mathrm{LCB}}^{(t+1)}(\exprime) = \mathcal{O}(\gamma^t \|C\|^t) + (1 + \ldots + \gamma^{t} C^{t})CR -\sqrt{\mathds{E} \Big[\Big( (1 + \ldots + \gamma^{t} C^{t}) (Q^{(0)}_{\theta^i}(\exprime) - CQ^{(0)}_{\theta^i}(\ex))\Big)^2\Big]}\\
                &\textbf{Shared Targets (Method 2):}\nonumber\\
                &Q_{\mathrm{LCB}}^{(t+1)}(\exprime) = \mathcal{O}(\gamma^t \|C\|^t) + (1 + \ldots + \gamma^{t} C^{t})C \arr - (1 + \ldots + \gamma^{t} C^{t}) \sqrt{\mathds{E}\Big[\Big(Q^{(0)}_{\theta^i}(\exprime) - C Q^{(0)}_{\theta^i}(\ex)\Big)^2\Big]}
            \end{align}
        where the square and square-root operations are applied element-wise.
        \footnote{Note that if $\gamma\|C\| \ge 1$, policy evaluation is liable to diverge in either setting. In our discussions, we avoid this degenerate case and assume $\gamma\|C\| < 1$.}
        \begin{proof}
            First we will establish some additional notation.

Consider an infinite-width NTK-parameterized \citep{jacot2018neural} $Q$-function and let $\q{t}{s,a}$ denote its value prediction for state-action pair $(s,a)$ after $t$ iterations of pessimistic policy evaluation. We will use the notation $\theta^{(t)}$ to refer to the parameters of this network after $t$ iterations, but for simplicity of notation, when the context is clear we will omit the use of $\theta^{(t)}$. Additionally, consider the linearization (Taylor expansion) of $Q$ about its initialization:
    \begin{align}
    Q_{\mathrm{lin}}(s,a) &:= \q{0}{s,a} + \nabla_\theta\q{0}{s,a} \cdot (\theta_{\mathrm{lin}} - \theta^{(0)})
    \end{align}
When performing pessimistic policy evaluation using linearized networks, we will use $\qlin{t}{s,a}$ to denote the value prediction of $Q_{\mathrm{lin}}$ after $t$ iterations, and will use $\theta_{\mathrm{lin}}^{(t)}$ to refer to its learned weights. Note that $\theta_{\mathrm{lin}}^{(0)} = 0$, and that $\forall (s,a), \q{0}{s,a} = \qlin{0}{s,a}$.

In \citet{lee2019wide} (section 2.4) it is shown that when training an infinitely wide neural network to perform regression using mean squared error, subject to technical conditions on the learning rate used, the predictions of the trained network are equivalent to if we had trained the linearized network instead. This means that after $t$ iterations of our policy evaluation procedure, $\forall (s,a), t, \qlin{t}{s,a} = \q{t}{s,a}$.
\emph{Hereafter we study the evolution of ensembles of linearized infinite-width networks, $Q_{\mathrm{lin}}$, across training iterations.}

As a reminder, $\mathcal{X}, \arr, \ex'$ denote data matrices containing $(s,a)$, $r$, and $(s',\pi(s'))$ appearing in the offline dataset $D$; i.e., the $k$-th transition $(s,a,r,s')$ in $D$ is represented by the $k$-th rows in $\mathcal{X},\arr,\ex'$. Additionally, we defined $C := \tzero(\exprime, \ex) \cdot \tzero(\mathcal{X}, \mathcal{X})^{-1}$.
Below, we will use notation such as $Q(\ex)$ to mean applying $Q$ to each row of $\ex$ and stacking the predictions into a vector. By $\tzero((s,a), \ex)$ we mean to treat $(s,a)$ as a row matrix and compute the $1 \times \vert D\vert$ kernel matrix.

\paragraph{Derivation for Independent Targets}
    When using Independent Targets, in iteration $t+1$ each network uses its own temporal difference (TD) targets,
        \begin{equation}
            \Why{t} = \arr + \gamma\qlin{t}{\exprime} \label{eq:target_independent}
        \end{equation}
    Using the equations in \citet{lee2019wide} (section 2.2, equations 9-10-11) we can derive,
        \begin{align}
            \qlin{t+1}{\ex} &= \Why{t}\\
            \forall s,a, \qlin{t+1}{s,a} &= \q{0}{s,a} + \Tzero{(s,a), \ex} \cdot \Tzero{\ex, \ex}^{-1} \cdot \left(\Why{t} - \q{0}{\ex}\right)\\
            \qlin{t+1}{\exprime} &= \q{0}{\exprime} + \Tzero{\exprime, \ex} \cdot \Tzero{\ex, \ex}^{-1} \cdot \left(\Why{t} - \q{0}{\ex}\right)
        \end{align}
    Plugging in the expression for the targets $\Why{t}$ and recursively expanding the expressions above we obtain,
        \begin{align}
            \qlin{t+1}{\exprime} &= \q{0}{\exprime} + \Tzero{\exprime, \ex} \cdot \Tzero{\ex, \ex}^{-1} \cdot \left(\Why{t} - \q{0}{\ex}\right)\\
            &= \q{0}{\exprime} + \Tzero{\exprime, \ex} \cdot \Tzero{\ex, \ex}^{-1} \cdot \left(\arr + \gamma\qlin{t}{\exprime} - \q{0}{\ex}\right)\\
            &= \q{0}{\exprime} + C \cdot \left(\arr + \gamma\qlin{t}{\exprime} - \q{0}{\ex}\right)\\
            &= \q{0}{\exprime} + C\arr - C\q{0}{\ex} + \gamma C\qlin{t}{\exprime}\\
            &= \ldots\\
            &= (1 + \ldots + \gamma^{t} C^{t}) \Big(
            \q{0}{\exprime} + C\arr - C\q{0}{\ex}\Big) + (\gamma C)^{t+1} \q{0}{\exprime} \label{eq:ind_closed_form}
        \end{align}
    Per our earlier discussion above, from Equation \ref{eq:ind_closed_form} we can conclude that,
        \begin{equation}
            \q{t+1}{\exprime} = \qlin{t+1}{\exprime} = (1 + \ldots + \gamma^{t} C^{t}) \Big(
            \q{0}{\exprime} + C\arr - C\q{0}{\ex}\Big) + (\gamma C)^{t+1} \q{0}{\exprime}
        \end{equation}
    We can now derive $Q_{\mathrm{LCB}}$ for the Independent Targets setting, by computing the expectation and variance of $\q{t+1}{\exprime}$ with respect to the initial weight distribution. Following \citet{lee2019wide} we obtain,
        \begin{align}
            \mathds{E}[\q{t+1}{\exprime}] &= (1 + \ldots + \gamma^{t} C^{t}) C \arr \\
            \mathrm{Var}[\q{t+1}{\exprime}] &= \mathds{E} \Bigg[\Bigg( \Big(1 + \ldots + \gamma^{t} C^{t}\Big) \Big(\q{0}{\exprime} - C\q{0}{\ex}\Big) + (\gamma C)^{t+1} \q{0}{\exprime} \Bigg)^2\Bigg]\\
            &= \mathds{E} \Bigg[\Bigg( \Big(1 + \ldots + \gamma^{t} C^{t}\Big) \Big(\q{0}{\exprime} - C\q{0}{\ex}\Big)\Bigg)^2\Bigg] + \mathcal{O}(\gamma^t \|C\|^t)
        \end{align}
    We thus have,
        \begin{tcolorbox}
            \begin{align}
                \qlcb{t+1}{\exprime} &= \mathds{E}[\q{t+1}{\exprime}] - \sqrt{\mathrm{Var}[\q{t+1}{\exprime}]} \label{eq:proof_part_1}\\
                &= \mathcal{O}(\gamma^t \|C\|^t) + (1 + \ldots + \gamma^{t} C^{t})CR -\sqrt{\mathds{E} \Big[\Big( (1 + \ldots + \gamma^{t} C^{t}) (Q^{(0)}_{\theta^i}(\exprime) - CQ^{(0)}_{\theta^i}(\ex))\Big)^2\Big]} \nonumber
            \end{align}
        \end{tcolorbox}
    where the square and square-root operators are applied element-wise.
    Note that if $\gamma\|C\| \ge 1$, policy evaluation is liable to diverge. Thus, in our discussions we avoid this degenerate case and assume $\gamma\|C\| < 1$, which makes the term $\mathcal{O}(\gamma^t \|C\|^t)$ negligible as $t \rightarrow \infty$.

\paragraph{Derivation for Shared LCB Targets}
    When using Shared LCB Targets, in iteration $t+1$ each network uses pessimistic temporal difference (TD) targets that are shared amongst ensemble members,
        \begin{align}
            \Why{t} &= \mathrm{LCB}\Big(\arr + \gamma\qlin{t}{\exprime}\Big) \\
            &= \arr + \mathrm{LCB}\Big(\gamma\qlin{t}{\exprime}\Big) \label{eq:target_shared_lcb}
        \end{align}
    Using the equations in \citet{lee2019wide} (section 2.2, equations 9-10-11) we can derive,
        \begin{align}
            \qlin{t+1}{\ex} &= \Why{t}\\
            \forall s,a, \qlin{t+1}{s,a} &= \q{0}{s,a} + \Tzero{(s,a), \ex} \cdot \Tzero{\ex, \ex}^{-1} \cdot \left(\Why{t} - \q{0}{\ex}\right)\\
            \qlin{t+1}{\exprime} &= \q{0}{\exprime} + \Tzero{\exprime, \ex} \cdot \Tzero{\ex, \ex}^{-1} \cdot \left(\Why{t} - \q{0}{\ex}\right)\\
            &= \q{0}{\exprime} + C \cdot \left(\Why{t} - \q{0}{\ex}\right)
        \end{align}
    Noting that in the Shared-LCB setting $\Why{t}$ is not a random variable, we can now compute the expectation and variance of $\qlin{t+1}{\exprime}$ with respect to the initial weight distribution,
        \begin{align}
            \mathds{E}[\qlin{t+1}{\exprime}] &= \mathds{E}\Big[\q{0}{\exprime} + C \cdot \left(\Why{t} - \q{0}{\ex}\right)\Big] = C\Why{t}\\
            \mathrm{Var}[\qlin{t+1}{\exprime}] &= \mathds{E}\Big[\Big(\q{0}{\exprime} - C \q{0}{\ex}\Big)^2\Big] = \text{same $\forall t+1 \ge 1$}
        \end{align}
    Let $A := \sqrt{\mathrm{Var}[\qlin{t+1}{\exprime}]}$. Given the above equations, we can recursively compute the closed form for $\Why{t}$ as follows,
        \begin{align}
            \Why{t} &= \arr + \mathrm{LCB}\Big(\gamma\qlin{t}{\exprime}\Big)\\
            &= \arr + \gamma \cdot \Big[\mathds{E}[\qlin{t}{\exprime}] - \mathrm{Var}[\qlin{t}{\exprime}]\Big]\\
            &= \arr + \gamma \cdot \Big[C \Why{t-1} - A\Big]\\
            &= \arr - \gamma A + \gamma C \Why{t-1}\\
            &= \ldots \\
            &= (1 + \ldots + \gamma^{t-1}C^{t-1})(\arr - \gamma A) + \gamma^{t}C^t\Why{0}\\
            &= (1 + \ldots + \gamma^{t-1}C^{t-1})(\arr - \gamma A) + \gamma^{t}C^t \arr + \gamma^{t+1}C^t \mathrm{LCB}(\q{0}{\exprime})
        \end{align}
    Based on this form, we can write,
        \begin{align}
            \mathds{E}[\qlin{t+1}{\exprime}] &= C\Why{t} \\
            &= (1 + \ldots + \gamma^{t}C^{t})C\arr - (\gamma C + \ldots + \gamma^{t}C^{t})A + \gamma^{t+1}C^{t+1} \mathrm{LCB}(\q{0}{\exprime})
        \end{align}
    Combining our derivations, and noting from our earlier discussion that $\forall (s,a), t, \qlin{t}{s,a} = \q{t}{s,a}$, we have,
        \begin{align}
            \qlcb{t+1}{\exprime} &= \mathds{E}[\q{t+1}{\exprime}] - \sqrt{\mathrm{Var}[\q{t+1}{\exprime}]} \\
            &= \mathds{E}[\qlin{t+1}{\exprime}] - \sqrt{\mathrm{Var}[\qlin{t+1}{\exprime}]} \\
            &= (1 + \ldots + \gamma^{t}C^{t})C\arr - (\gamma C + \ldots + \gamma^{t}C^{t})A + \gamma^{t+1}C^{t+1} \mathrm{LCB}(\q{0}{\exprime}) - A \\
            &= (1 + \ldots + \gamma^{t}C^{t})C\arr - (1 + \ldots + \gamma^{t}C^{t})A + \gamma^{t+1}C^{t+1} \mathrm{LCB}(\q{0}{\exprime})
        \end{align}
    Thus, we have our results,
        \begin{tcolorbox}
            \begin{align}
                \qlcb{t+1}{\exprime} &= \mathcal{O}(\gamma^t \|C\|^t) + (1 + \ldots + \gamma^{t} C^{t})CR - (1 + \ldots + \gamma^{t}C^{t}) \sqrt{\mathds{E}\Big[\Big(Q^{(0)}_{\theta^i}(\exprime) - C Q^{(0)}_{\theta^i}(\ex)\Big)^2\Big]} \label{eq:proof_part_2}
            \end{align}
        \end{tcolorbox}
    where the square and square-root operators are applied element-wise.
    Note that if $\gamma\|C\| \ge 1$, policy evaluation is liable to diverge. Thus, in our discussions we avoid this degenerate case and assume $\gamma\|C\| < 1$, which makes the term $\mathcal{O}(\gamma^t \|C\|^t)$ negligible as $t \rightarrow \infty$.
    
    The derivations of Equations \ref{eq:proof_part_1} and \ref{eq:proof_part_2} conclude our proof.

        \end{proof}
    \end{theorem}

\section{A Pedagogical Toy MDP}
\label{app:old_toy_mdps}
    \begin{figure}[H]
        \centering
        \includegraphics[width=\textwidth]{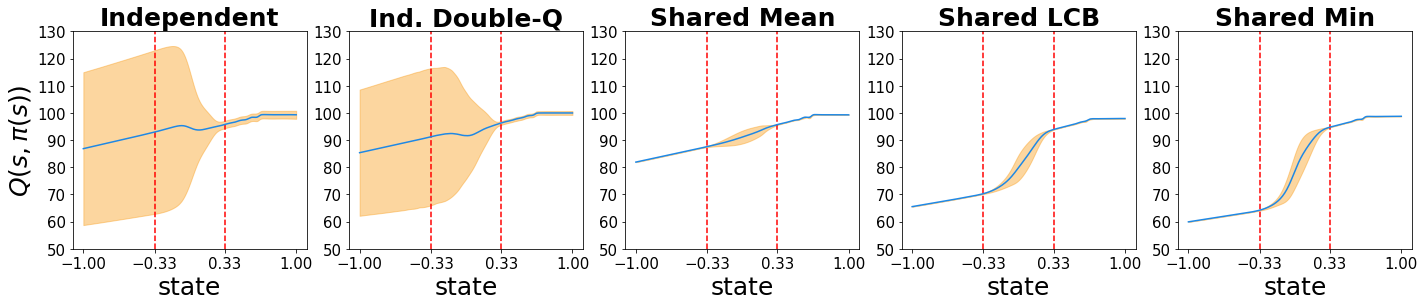}
        \caption{\small{Verifying theoretical predictions on the toy Continuous Chain MDP. The marked interval $[-0.33, 0.33]$ denotes the region of state-space with no data. As anticipated by Theorem \ref{theorem:modulation}, when the $Q$-value functions are trained independently, the derived uncertainties capture the interaction between the available data, the structure of the MDP, and the policy being evaluated.
        }}
        \label{fig:verify_theory}
    \end{figure}
    
    \subsection{Qualitative Evaluation of Obtained Uncertainties}
    \label{sec:toy_exp}
    
    In this section we continue to study the implications of our theoretical results in Section \ref{sec:independence_matters} by constructing a pedagogical toy MDP.
    Our simple toy MDP allows us to follow the idealized setting of the presented theorem more closely, and allows for visualization of uncertainties obtained through different ensembling approaches.
    
    \paragraph{Continuous Chain MDP} The MDP we consider has state-space $\mathcal{S} = [-1, 1]$, action space $\mathcal{A} \in \mathds{R}$, deterministic transition dynamics $s' = s + a$ clipped to remain inside $\mathcal{S}$, and reward function $r(s,a) = \mathds{1}[s' \in [0.75, 1]]$.
    
    \paragraph{Data Collection \& Evaluation Policy} The offline dataset we generate consists of 40 episodes, each of length 30. At the beginning of each episode we initialize at a random state $s \in \mathcal{S}$. In each step we take a random action sampled from $\text{Unif}(-0.3, 0.3)$, and record all transitions $(s, a, r, s')$. For evaluating the uncertainties obtained from different approaches, we create regions of missing data by removing all transitions where $s$ or $s'$ fall in the range $[-0.33,0.33]$. The policy used for policy evaluation with the different ensembling approaches is $\forall s, \pi(s) = 0.1$.
    
    \paragraph{Optimal Desired Form of Uncertainty} Note that the evaluation policy $\pi(s) = 0.1$ is always moving towards the positive direction, and there is lack of data for states in the interval $[-0.33, 0.33]$. Hence, what we would expect is that in the region $[0.33, 1]$ there should not be a significant amount of uncertainty, while in the region $[-1, 0.33]$ there should be significantly more uncertainty about the Q-values of $\pi$, because the policy will be passing through $[-0.33, 0.33]$ where there is no data. Furthermore, as we move towards the negative axis, we would expect that in the region $[-0.33, 0.33]$ the uncertainty would gradually increase, while in the region $[-1, -0.33]$ the uncertainty would not continue to increase.
    
    
    \paragraph{Results} We visualize and compare the uncertainties obtained when the targets in the pessimistic policy evaluation procedure are computed as:
    \begin{itemize}
        \item \textbf{Independent Targets (as used in MSG):} $y^i = r + \gamma \cdot Q_{\theta^i}(s', \pi(s'))$
        \item \textbf{Independent Double-Q Targets:} $y^i = r + \gamma \cdot \mathrm{min} \Big[ Q^1_{\theta^i}(s', \pi(s')), Q^2_{\theta^i}(s', \pi(s'))\Big]$
        \item \textbf{Shared Mean Targets:} $y = r + \gamma \cdot \mathrm{mean} \Big[Q_{\theta^i}(s', \pi(s'))\Big]$
        \item \textbf{Shared LCB Targets:} $y = r + \gamma \cdot \Bigg[\mathrm{mean} \Big[Q_{\theta^i}(s', \pi(s'))\Big] - 2 \cdot \mathrm{std} \Big[Q_{\theta^i}(s', \pi(s'))\Big]\Bigg]$
        \item \textbf{Shared Min Targets:} $y = r + \gamma \cdot \mathrm{min} \Big[Q_{\theta^i}(s', \pi(s'))\Big]$
    \end{itemize}
    
    
    Note that \texttt{Independent Double-Q} is still an independent ensemble, where each ensemble member has an architecture containing a min-pooling on top of two subnetworks.
    
    In Figure \ref{fig:verify_theory} we plot the mean and two standard deviations of the $Q$-values predicted by the ensemble for the policy we evaluated, $\pi(s) = 0.1$ (additional experimental details presented in Appendix \ref{app:toy_details}). The first striking observation is that, \texttt{Independent} targets (as used in MSG) effectively match our desired form of uncertainty: states that under the evaluation policy $\pi(s) = 0.1$ would lead to regions with little data have wider uncertainties than states that do not. A second observation is that, in this toy construction, \texttt{Shared LCB} and \texttt{Shared Min} provide a seemingly good approximation to the lower-bound of \texttt{Independent} predictions. However, our theoretical results show that shared targets have critical failure cases. Furthermore, our empirical results on challenging benchmark domains (Section \ref{sec:d4rl_antmaze}) demonstrate that
    \texttt{Shared LCB} and \texttt{Shared Min} targets completely fail to train successful policies, despite their implementation differing from \texttt{Independent} targets in only 2 lines of code.

    \subsection{NTK vs. Maximal Parameterization}
        \label{app:additional_toy}
        
        The toy experiment presented in section \ref{sec:toy_exp} uses a single-hidden layer finite-width neural network architecture with \texttt{tanh} activations. The networks use the ``standard weight parameterization" (i.e. the weight parameterization used in practice) as opposed to the NTK parameterization~\citep{novak2019neural}, and we optimize the networks using the Adam optimizer \citep{kingma2014adam}. While this setup is close to the practical setting and demonstrates the relevance of independent ensembles for the practical setting, an important question posed by our reviewers is how close these results are to the theoretical predictions presented in \ref{theorem:modulation}. To answer this question, we present the following set of results.
        
        Using the identical MDP and offline data as before, we implement 1 hidden layer neural networks with \texttt{erf} non-lineartiy. The networks are implemented using the Neural Tangents library \citep{novak2019neural}, and use the NTK parameterization. The networks in the ensemble are optimized using full-batch gradient descent with learning rate 1 for 500 steps of Fitted $Q$-Evaluation (FQE) \cite{fonteneau2013batch}, where in each FQE step the networks are updated for 1000 gradient steps. We vary the width of the networks from 32 to 32768 in increments of a factor of $4$, plotting the mean and standard deviation of the network predictions. The ensemble size is set to $N = 16$, except for width $32768$ where $N = 4$.
        
        We compare the results for finite-width networks to computing
        results for
        the infinite-width setting in closed form (using Theorem \ref{theorem:modulation}). Using the Neural Tangents library \citep{novak2019neural} we obtained the NTK for the architecture described in the previous paragraph (1 hidden layer with \texttt{erf} non-linearity). We found that the matrix inversion required in our equations results in numerical errors. Hence, we make the modification $\Theta(\ex, \ex) \leftarrow \Theta(\ex, \ex) + \texttt{1e-3} \cdot \text{I}$.
        
\begin{figure}[H]
            \centering
            \includegraphics[width=\textwidth]{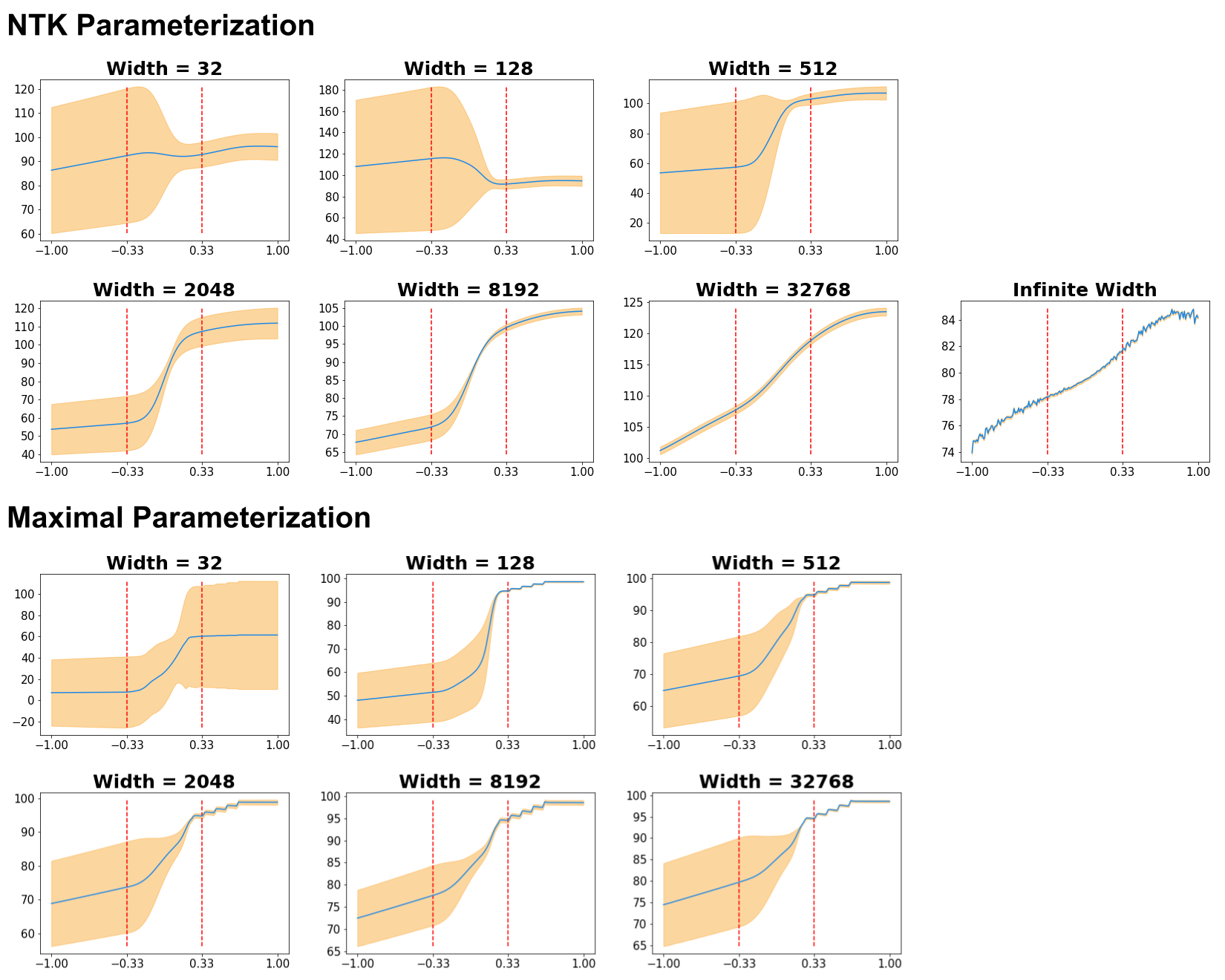}
            \caption{Comparing results of finite-width networks to closed form equations derived in Theorem \ref{theorem:modulation}. In the NTK parameterization, as width $\rightarrow \infty$, the structure of the variances collapse and resemble the infinite-width closed-form results. We believe this is due to infinite-width networks under the NTK regime not being able to learn features \citep{yang2020feature}. Supporting this hypothesis, we observe that networks parameterized by the Maximal Parameterization of \citet{yang2020feature} maintain the desired uncertainty structure as the width of the networks grows larger.}
            \label{fig:inf_width_results}
        \end{figure}

        Figure \ref{fig:inf_width_results} presents our results. As the width of the networks grow larger, the shape of the uncertainties becomes more similar to our closed-form equations (i.e. the variances become very small). While we do not have a rigorous explanation for why finite-width networks exhibit intuitively more desirable behaviors, we present below a strong hypothesis backed by empirical evidence. We believe rigorously answering this question is an incredibly interesting avenue for future work.
        
        Hypothesis: Infinite-width networks in the NTK parameterization do not learn data-dependent features \citep{yang2020feature}.
        \citet{yang2020feature} present a different approach for parameterizing infinite-width networks called the ``Maximal Parameterization", which enables inifinite-width networks to learn data-dependent features. We perform the same experiment as above, by replacing the NTK-parameterized networks with Maximal Parameterizations. Figure \ref{fig:inf_width_results} presents our empirical results for network widths from 32 to 32768. Excitingly, we observe that with Maximial Parameterization, even our widest networks recover the intuitively desired form of uncertainty described in section \ref{sec:toy_exp}! The solutions of these networks also appear much more accurate, particularly on the right hand side of the plot; we can observe the correct stepped structure of the $Q-values$ up until the interval $[0.75, 1]$ where the policy always receives a reward of $1$. Each step appears to be approximately $0.1$ in width, which is size of the action of the policy being evaluated, $\forall s, \pi(s) = 0.1$.
        
    \subsection{Additional Implementation Details for Figure \ref{fig:verify_theory}}
    \label{app:toy_details}
        To evaluate the quality of uncertainties obtained from different $Q$-function ensembling approaches, we create $N=64$ $Q$-function networks, each being a one hidden layer neural network with hidden dimension 512 and \texttt{tanh} activation. The initial weight distribution is a fan-in truncated normal distribution with scale 10.0, and the initial bias distribution is fan-in truncated normal distribution with scale 0.05. We did not find results with other activation functions and choices of initial weight and bias distribution to be qualitatively different. We use discount $\gamma = 0.99$ and the networks are optimized using the Adam \citep{kingma2014adam} optimizer with learning rate \texttt{1e-4}.
        In each iteration, we first compute the TD targets using the desired approach (e.g. independent vs. shared targets) and then fit the $Q$-functions to their respective targets with 2000 steps of full batch gradient descent. We train the networks for 1000 such iterations (for a total of $2000 \times 1000$ gradient steps). Note that we do not use target networks.
        Given the small size of networks and data, these experiments can be done within a few minutes using a single GPU in a \texttt{Google Colaboratory} notebook which we have included in our opensourced codebase.

\section{Statistical Model}
    \begin{figure}[H]
        \centering
        \begin{subfigure}{0.49\textwidth}
            \centering
            \begin{tikzpicture}

  \node[latent]                               (Q) {$Q$};
  \node[obs, right=of Q, xshift=0cm] (arch) {arch};
  \node[obs, above=of Q] (pol) {$\pi$};
  \node[obs, right=of Q, yshift=1.7cm]  (D) {$D$};
  \node[obs, left=of Q]  (alg) {alg};

  \edge {arch, pol, D, alg} {Q} ; %

\end{tikzpicture}
            \caption{Q-function generative process: the graphical model above represents the induced distribution over Q-functions when conditioning on a particular policy evaluation algorithm, policy, offline dataset, and Q-function network architecture.}
            \label{fig:pgm0}
        \end{subfigure}
        \centering
        \begin{subfigure}{0.49\textwidth}
            \centering
            \begin{tikzpicture}

  \node[obs]                               (Q) {$Q$};
  \node[latent, right=of Q, xshift=0cm] (arch) {arch};
  \node[obs, right=of arch, xshift=0cm] (prior) {prior};
  \node[obs, above=of Q] (pol) {$\pi$};
  \node[obs, right=of Q, yshift=1.7cm]  (D) {$D$};
  \node[obs, left=of Q]  (alg) {alg};

  \edge {arch, pol, D, alg} {Q} ; %
  \edge {prior} {arch} ; %
  
  \plate {polQ} {(pol)(Q)} {$N$} ;

\end{tikzpicture}
            \caption{An example of an interesting extension}
            \label{fig:pgm1}
        \end{subfigure}
    \end{figure}
    
    An interesting question posed by reviewers of our work was ``[W]hatever formal reasoning system we'd like to use, what is the ideal answer, given access to arbitrary computational resources, so that approximations are unnecessary? I.e., how do we quantify our uncertainty about the MDP and value function before seeing data, and how do we quantify it after seeing data?"
    
    It is important to begin by clarifying what is the mathematical object we are trying to obtain uncertainties over. In this work, we do not quantify uncertainties about any aspects of the MDP itself (although this is an interesting question which comes up in model-based methods as well as other settings such as Meta-Learning \citep{yu2019meta, ghasemipour2019smile}). Our goal in this work is to directly estimate $Q^\pi(s, a) = r(s,a) + \gamma \cdot \mathds{E}_{s' \sim MDP, a' \sim \pi}[Q(s', a')]$, for $a \sim \pi(s)$, and to obtain uncertainties about $Q^\pi(s,a)$.
    
    Let $Q(s,a)$ be a predictor $\mathcal{S} \times \mathcal{A} \rightarrow \mathds{R}$ that needs to be evaluated on -- and hopefully generalize well to -- $(s,a) \notin D$, where $D$ is the offline dataset. When we choose to represent $Q(s,a)$ using neural networks, Gaussian Processes, or K-nearest-neighbours, we are not just making approximations for computational reasons, but are actually choosing a function class which we believe will generalize well to unseen $(s,a)$.
    
    One practical example of learning $Q$-functions is to use Fitted $Q$-Evaluation (FQE) \citep{fonteneau2013batch} on the provided data using gradient descent with a desired neural network architecutre. Due to the random weight initialization, this procedure induces a distribution on the $Q$-functions which is captured by the probabilistic graphical model (PGM) in Figure \ref{fig:pgm0}. In other words, by conditionining on the policy, data, architecture, and policy evaluation algorithm, we are imposing a belief over $Q$-functions. Note that this is effectively the same justification as using ensembles in supervised deep learning, where ensembles are state-of-the-art for accuracy and calibration \citep{ovadia2019can}. For the sake of theoretical analysis (Section \ref{sec:independence_matters}), we studied this belief distribution under the infinite-width NTK~\citep{jacot2018neural,lee2019wide} network setting, in which case the distribution over $Q$-functions is a Gaussian Process.
    
    The focus of this work is to ask the question: ``Under this imposed belief, what should the policy update be?". Our proposed answer is to optimize the policy with respect to the lower-confidence bound of our beliefs:
    In the standard actor-critic setup, the policy optimization objective takes the form $\max_\pi\mathds{E}_{d(s)}[Q(s,\pi(s))]$, where $d(s)$ is some distribution over states (e.g. the initial state distribution, a uniform distribution over states in the offline dataset $D$, etc.).
    For the pessimistic offline RL settting, our proposed policy optimization objective takes the form $\max_\pi\text{LCB}\Big(\mathds{E}_{d(s)}[Q(s,\pi(s))]\Big)$. In MSG, for practical reasons, we convert this to $\max_\pi\mathds{E}_{d(s)}\Big[\text{LCB}\Big(Q(s,\pi(s))\Big)\Big]$ (which is a lower-bound of the latter objective).
    
    The graphical model in Figure \ref{fig:pgm1} also highlights an example of interesting future directions: Consider an offline RL setup where we keep track of the various policies generating the data and their $Q$-functions (approximated through Monte-Carlo estimation). Then, we can impose a prior distribution over architectures, and using the available $Q$-functions, infer a posterior distribution over architectures (for a probabilistic interpretation, we could treat the output of a $Q$-function as the mean of a standard normal distribution). Subsequently, given a new policy, we can learn its $Q$-function under the posterior neural network architecture distribution.

\end{document}